\documentclass[10pt,journal,compsoc]{IEEEtran}
\usepackage{amsmath,amsfonts}
\usepackage{algorithmic}
\usepackage{algorithm}
\usepackage{array}
\usepackage[caption=false,font=normalsize,labelfont=sf,textfont=sf]{subfig}
\usepackage{textcomp}
\usepackage{stfloats}
\usepackage{url}
\usepackage{verbatim}
\usepackage{graphicx}
\usepackage{cite}

\usepackage{amsfonts}
\usepackage{amsmath}
\usepackage{multirow}
\usepackage{booktabs}
\usepackage{array}
\usepackage{footnote}

\usepackage[colorlinks = true,
            linkcolor = blue,
            urlcolor  = blue,
            citecolor = blue,
            anchorcolor = blue]{hyperref}

\begin{document}

\title{FALCON: Feature-Label Constrained Graph Net Collapse for Memory Efficient
GNNs}

\author{Christopher Adnel and Islem Rekik, \IEEEmembership{Member, IEEE}
\thanks{C. Adnel and I. Rekik are affiliated with BASIRA Lab, Imperial-X and Department of Computing, Imperial College London, London, UK. Corresponding author: Islem Rekik, Email: i.rekik@imperial.ac.uk, \url{https://basira-lab.com} and \url{https://ix.imperial.ac.uk/}}.}

\markboth{IEEE Journal 2023}%
{Shell \MakeLowercase{\textit{et al.}}: A Sample Article Using IEEEtran.cls for IEEE Journals}


\maketitle

\begin{abstract}
Graph Neural Network (GNN) ushered in a new era of machine learning with interconnected datasets. While traditional neural networks can only be trained on independent samples, GNN allows for the inclusion of inter-sample interactions in the training process. This gain, however, incurs additional memory cost, rendering most GNNs unscalable for real-world applications involving vast and complicated networks with tens of millions of nodes (e.g., social circles, web graphs, and brain graphs). This means that storing the graph in the main memory can be difficult, let alone training the GNN model with significantly less GPU memory. While much of the recent literature has focused on either mini-batching GNN methods or quantization, graph reduction methods remain largely scarce. Furthermore, present graph reduction approaches have several drawbacks. First, most graph reduction focuses only on the inference stage (e.g., condensation, pruning, and distillation) and requires full graph GNN training, which does not reduce training memory footprint. Second, many methods focus solely on the graph's structural aspect, ignoring the initial population feature-label distribution, resulting in a skewed post-reduction label distribution. Here, we propose a Feature-Label COnstrained graph Net collapse, FALCON, to address these limitations. Our three core contributions lie in (i) designing FALCON, a topology-aware graph reduction technique that preserves feature-label distribution by introducing a K-Means clustering with a novel dimension-normalized Euclidean distance; (ii) implementation of FALCON with other state-of-the-art (SOTA) memory reduction methods (i.e., mini batched GNN and quantization) for further memory reduction; (iii) extensive benchmarking and ablation studies against SOTA methods to evaluate FALCON memory reduction. Our comprehensive results show that FALCON can significantly collapse various public datasets (e.g., PPI and Flickr to as low as 34\% of the total nodes) while keeping equal prediction quality across GNN models. Our FALCON code is available at \url{https://github.com/basiralab/FALCON}.
\end{abstract}


\section{Introduction}
\IEEEPARstart{T}{he} advent of Graph Neural Networks (GNNs) has fostered a new neural network learning paradigm where inter-sample dependencies can be taken into account in addition to their individual feature. Consequently, this opens up neural networks to many new practical applications that involve graph datasets such as network neuroscience \cite{basira-neuroscience}, retail recommendations \cite{recommender}, fraud detections \cite{wu2023energybased}, and drug discovery \cite{life_science}. However, \emph{scalability} remains one of the main barriers in GNNs as most of these practical applications usually involve large and complex graph datasets. During our experiments, out-of-memory (OOM) errors were also very frequent when training GNN models (e.g., GCN), even on relatively smaller datasets (e.g., Flickr).

Furthermore, most GNN models such as Graph Convolutional Network (GCN) \cite{GCN}, Graph Attention Network (GAT) \cite{GAT}, and Neural Message Passing Scheme (NPMS) \cite{NPMS} couple the feature aggregation with feature transformation on each layer. This coupling fosters data point interconnectivity, making mini-batching complex. Consequently, early GNN training is mostly done with full-batched gradient descent, which does not scale well \cite{cluster-gcn}. While mini-batched stochastic gradient descent can still be done by including one-hop neighborhood nodes to the mini-batch for every GNN layer, this will result in exponential time complexity relative to the number of layers \cite{cluster-gcn}.

\textbf{Related Works.} Many recent works aim to make these GNN methods more scalable through clever mini-batching schemes considering node interconnectivity. One of the early mini-batching methods, Cluster-GCN \cite{cluster-gcn}, uses graph clustering algorithms such as METIS \cite{METIS} to partition the graph into subgraphs, formed by minimizing the inter-cluster edges while maximizing the within-cluster edges. Mini-batch can be composed of a combination of these clusters, and out-of-batch edges are neglected. GNN AutoScale \cite{GNNAutoScale:} is an improvement of Cluster-GCN by using stale out-of-batch node embeddings (historical embeddings) to compensate for the discarded out-of-batch edges. Similarly, LMC-GNN \cite{LMC-GNN} is also based on GNN AutoScale but utilizes a convex combination of historical embeddings and incomplete up-to-date messages to compensate for the out-of-batch edges. 

Interestingly, some recent works in this stream took a different path in constructing the mini-batch. SGC \cite{SGC} and SIGN \cite{frasca2020sign} simply decouple the feature aggregation and feature transformation phase. Under the assumption that the feature aggregation is fixed, SIGN and SGC pre-compute the aggregation and leave only feature transformation for MLP with independent data points.

\begin{figure}[t]
\centering
\includegraphics[width=0.9\columnwidth]{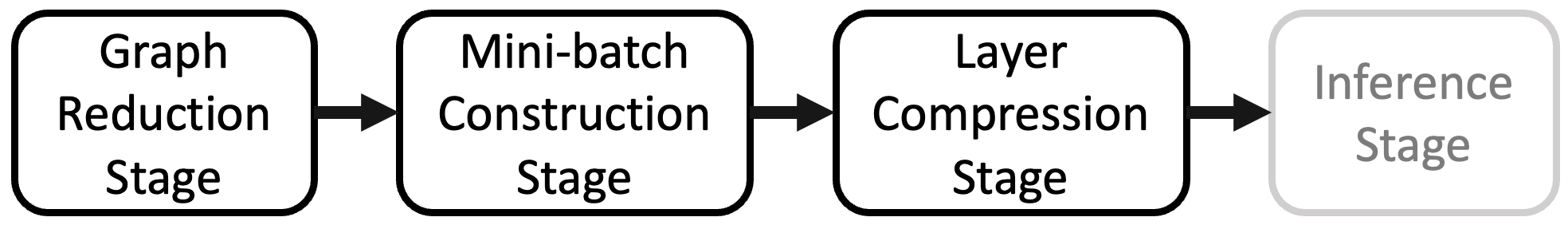}
\caption{Memory efficient GNN framework structure.}
\label{training_stages}
\end{figure}

Another stream of work in memory-efficient GNNs is quantization methods, which store each layer's data in a low-precision format to reduce memory usage. One of the most recent works along this direction, EXACT \cite{liu2022exact} applies the activation compression method originally developed for Convolutional Neural Network, ActNN \cite{chen2021actnn}.

Lastly, a relatively under-explored stream and the main focus of this present work is graph reduction methods. Such methods aim to find a compressed representation of the training graph which serves as a memory-efficient alternative to the original graph. Since graph reduction methods only compacts the training graph, they are independent from the types of GNN models being used and are very versatile to be applied in most GNN training pipelines. Additionally, most graph reduction methods such as \cite{scalablegraphreduction} and \cite{huang2021scaling} can be considered as a pre-processing task which is only performed once at the beginning of the training pipeline.

Some works in the graph reduction stream merge neighbouring nodes based on their label similarity (i.e., KL divergence) \cite{scalablegraphreduction} while another utilizes graph coarsening algorithms \cite{huang2021scaling} to reduce the training graph. In our recent work \cite{cqsign}, we experimented with various graph centrality measures to quantify individual nodes importance and drive the contraction. However, most of these existing graph reduction methods only consider the graph structural information and do not consider node features and labels, which implies that the original label and feature distribution might not be preserved. 

There are other methods relating to graph reduction, such as condensation \cite{condensation,condensation2} and pruning \cite{pruning}. However, these methods are intended for the inference stage and do not reduce training memory usage as the reduced graph is learned through full graph GNN training.

In this paper, we propose a novel graph-collapsing technique based on both the graph topological characteristics and the feature and label distribution across nodes. We name our method \textbf{FALCON: FeAture-Label COnstrained graph Net collapse} and further improve over \cite{cqsign}, our original work where we utilize graph centrality measure to rank each node based on its topological importance. However, improving from a pure topological approach, we also consider the nodes feature and label diversity to minimize any biases or imbalance introduced by the graph collapse. These feature embeddings and labels are used to guide the graph collapse to preserve the original feature and label diversity. 

Additionally, we also apply FALCON to the four-stage memory-efficient GNN framework introduced by our seed paper \cite{cqsign} along with various other state-of-the-art (SOTA) methods for mini-batching and quantization as visualized in Figure~\ref{training_stages}. However, since we are focusing on GNN training, we only work with the first three stages (i.e., graph reduction, mini-batch construction, and layer compression). 

The compelling apects of our work are summarized as follows:

\begin{itemize}
    \item FALCON presents the first GNN training-focused graph reduction method that utilizes both graph topological information and nodes feature and label data. FALCON achieves this through diversity-constrained graph centrality collapse which preserves feature-label distribution.
    \item FALCON has been extensively benchmarked on four public datasets from various fields, and we found all benchmark datasets to be highly collapsible, indicating its potential on wide range of datasets.
    \item Being part of a pre-processing stage, FACLON is highly compatible with many existing memory-efficient GNN frameworks. Furthermore, we propose a memory-efficient GNN framework incorporating FALCON and show the holistic impact of combining FALCON with other SOTA methods.
\end{itemize}

\section{Background}

\textbf{Problem statement.} Let $\mathcal{G} = \{\mathcal{V},\mathcal{E},X,Y\}$ denote a graph with $\mathcal{V}$ as the set of nodes or samples, $\mathcal{E}$ as the set of edges, $X \in \mathbb{R}^{|\mathcal{V}| \times F} $ as the feature matrix, and $Y \in \mathbb{R}^{|\mathcal{V}| \times L}$ as the label matrix. The objective of FALCON is to find a collapsed version of $\mathcal{G}$, namely $\mathcal{G'} = \{\mathcal{V}',\mathcal{E}',X',Y'\}$ such that $\mathcal{G}'$ fits within the defined memory budget $\Psi$ and GNN model $f_{\theta}$ trained on $\mathcal{G}$ will be approximately similar to the one trained on $\mathcal{G}'$. Formally, this is written as Equation~\ref{eqn:falcon}. Here, we mainly work with node classification tasks with unweighted and undirected graphs.
\begin{equation}\label{eqn:falcon}
\resizebox{0.9\hsize}{!}{
    $\mathbf{FALCON}(\mathcal{G}) = \mathcal{G}' \; s.t. \;  f_{\theta}(\mathcal{G}) \approx f_{\theta}(\mathcal{G}'), |f_{\theta}(\mathcal{G}')| < \Psi < |f_{\theta}(\mathcal{G})|$
    }
\end{equation}

Where $|f_{\theta}(\mathcal{G})|$ denotes the required memory budget to learn the mapping function $f_{\theta}$ on $\mathcal{G}$.

\textbf{Graph Convolution Network Model.} One of the earliest and least complex GNN models, GCN \cite{GCN} has layers composed of two phases, namely feature aggregation and feature transformation. While GCN feature aggregation uses a normalized adjacency matrix $\tilde{A}$ and does not involve any learnable parameters, it is coupled with the feature transformation phase due to the non-linear activation (e.g., ReLU) between each layer. Equation~\ref{eqn:GCN_main} shows a two-layer GCN network equation as follows:
\begin{equation}\label{eqn:GCN_main}
    \hat{Y} = \sigma(\tilde{A}\sigma(\tilde{A}X^{0}\theta^{1})\theta^{2})
\end{equation}
\begin{equation}\label{eqn:adjacency_normalization1}
\resizebox{0.9\hsize}{!}{$
    \tilde{A} = D^{-1/2} (A+I) D^{-1/2} \;\;\;\;
    D = 1 + \sum_j A_{ij}$
    }
\end{equation}

Where $A$ denotes the adjacency matrix, $\theta^{l}$ denotes the weights of the $l$\textsuperscript{th} layer, and $D$ denotes the diagonal matrix composed of node degrees.

This means that mini-batched training using GCN is not straightforward. We must include an additional one-hop neighborhood for each GCN layer, leading to an exponential time complexity w.r.t the number of layers. Hence, GPU memory footprint is a substantial obstacle in GCN training.

\section{Proposed FALCON}
\begin{figure*}[t]
\centering
\includegraphics[width=2.0\columnwidth]{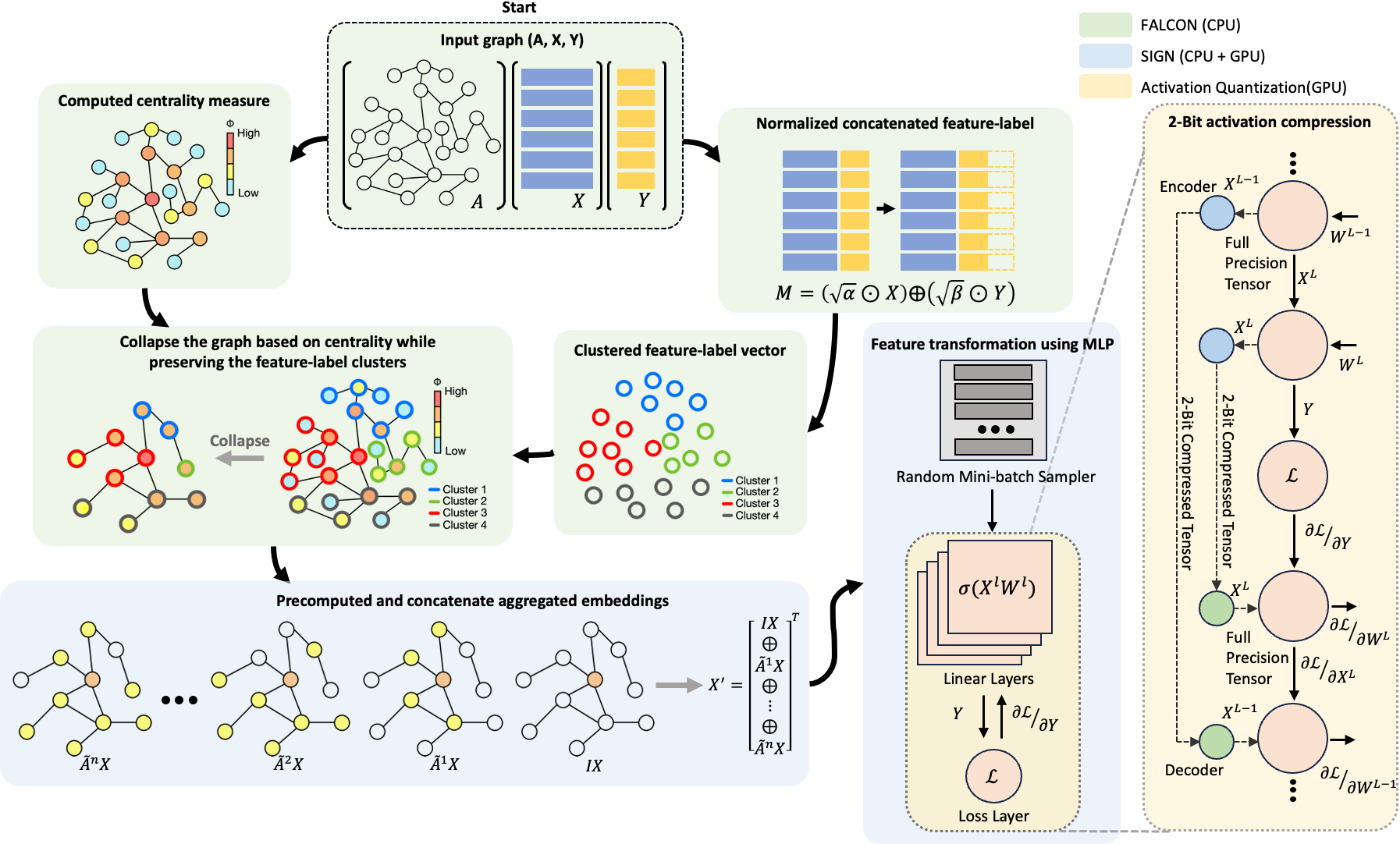}
\caption{FALCON-QSIGN flowchart overviews the proposed FALCON applied on activation quantized SIGN model \cite{frasca2020sign, chen2021actnn}. \textbf{FALCON.} From the starting block, FALCON diverges into 2 paths, where the left path is the centrality computation, and the right path is clustering using a normalized feature-label tensor. Afterward, the path converges, and we distribute the centrality-based collapse across each cluster to maintain the feature-label distribution. \textbf{SIGN.} In SIGN, we use the collapsed graph to compute the aggregated features and concatenate them to form a new feature tensor. Next, we pass the new feature tensor to an MLP. \textbf{Activation Quantization.} Here, we leverage activation quantization to the MLP by compressing the cached features into 2-bit precision.}
\label{main_figure}
\end{figure*}

This section details FALCON\footnote{\url{https://github.com/basiralab/FALCON}}. As shown in Figure~\ref{main_figure}, FALCON computation consists of 2 paths: centrality computation to encode topological awareness, and preserving the feature-label distribution by normalized clustering.

\begin{figure}[t]
\centering
\includegraphics[width=1\columnwidth]{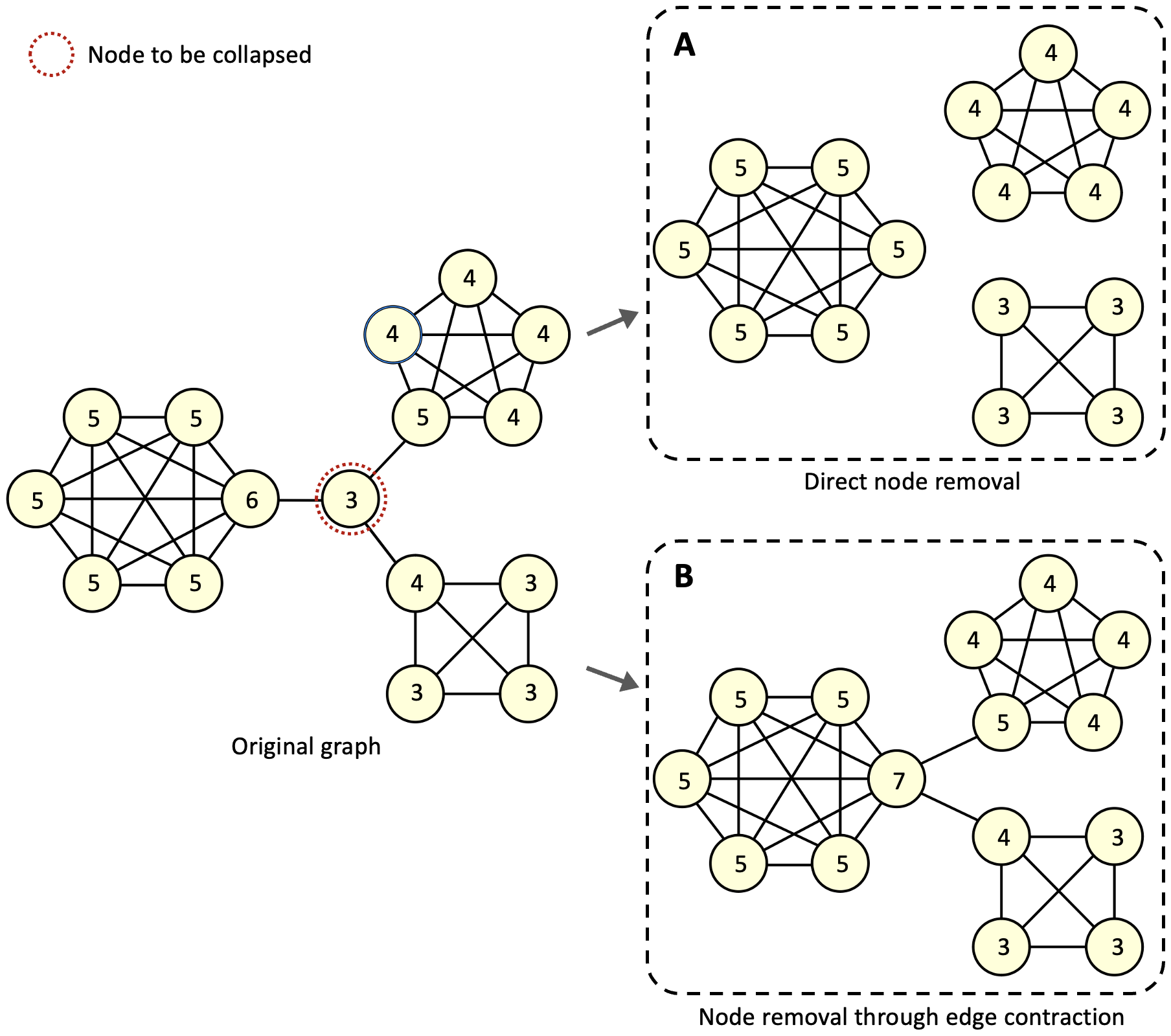}
\caption{Example of graph fragmentation caused by direct node removal (\textbf{A}), while removal through edge contraction maintains graph connectivity (\textbf{B}). The node number depicts the centrality measure (degree centrality).}
\label{edge_contraction}
\end{figure}

\textbf{Centrality-Based Graph Collapse.} Centrality measures are classical yet powerful tools to quantify node importance in the graph structure. Nodes deemed important usually mean they serve as a hub for others to communicate. Hence, centrality is usually defined based on factors relating to this "hubness" (e.g., the number of connections and shortest distance to every other node). Based on this, \cite{cqsign} contracts the graph starting from the least important node by merging it with its adjacent hub (the most important node within its neighborhood). In other words, they merge the weakest nodes iteratively with their strongest neighbors until the defined node budget $\Psi$ is fulfilled. Edge collapse is done for each merging to maintain the graph connectivity and avoid fragmenting the graph as depicted in Figure~\ref{edge_contraction}. Although pure centrality-based collapse does not guarantee the preservation of the original feature-label distribution, it remains as a promising metric to quantify node importance. Hence, we experimented with the following measures:

\begin{itemize}
\item \textbf{Degree Centrality (DC).} DC is the simplest centrality based only on the node's degree (a higher degree incurs higher centrality). DC can be computed using $DC(v_i)=\sum_{i \not= j} A_{ij}$ with $A$ as the graph adjacency matrix. Computation of DC is relatively fast as we only need to go through the set of edges $\mathcal{E}$ in linear time $O(|\mathcal{E}|)$.

\item \textbf{Betweenness Centrality (BC).} BC measures the shortest paths between any pair of nodes that crosses the node (more crossing incurs higher centrality). The computation of exact BC can be quite expensive with $O(|\mathcal{V}|.|\mathcal{E}|)$ time complexity \cite{fast_betweenness_centrality}. Fortunately, BC can be approximated by random sampling a small set of nodes $\mathcal{V}_{k}$ and computing BC based on $\mathcal{V}_{k}$. This leads to an upper bound time complexity of $O(|\mathcal{V}_{k}|.|\mathcal{E}|)$ \cite{CentralityTimeComplexity}. Formally, BC is defined as $BC(v_i)=\sum_{i \not= j \not= k} \frac{\sigma_{jk}(v_i)}{\sigma_{jk}}$ where $\sigma_{jk}(v_i)$ is the number of closest paths between $v_j$ and $v_k$ passing through $v_i$ and $\sigma_{jk}$ the total number of closest paths between $v_j$ and $v_k$.

\item \textbf{Closeness Centrality (CC).} CC measures how close a node is to every other node in the graph. This will naturally be computationally costly as it involves computing the shortest distance between all pairs of nodes. For sparse graphs, CC has $O(|\mathcal{V}|.|\mathcal{E}|)$ time complexity \cite{CentralityTimeComplexity}, making it one of the slowest centrality to compute. Formally, CC is defined as $CC(v_i) = \frac{|\mathcal{V}|-1}{\sum_{i \not= j} l_{ij}}$ where $l_{ij}$ is the closest distance between $v_i$ and $v_j$. 

\item \textbf{PageRank Centrality (PR).} Originally developed by Google \cite{pagerank} to rank websites for their search engine, PR measures a node centrality by taking into account the quality of the neighboring nodes and the probability of a random walker from that node visiting a node of interest. PR centrality is formally defined as $PR(v_i)=(I-\alpha A D^{-1})^{-1}\mathbf{1}_i$, with $D$ as diagonal matrix of node degrees and $\alpha$ is a damping factor usually set to 0.85. PR can be computed with $O(|\mathcal{V}|+|\mathcal{E}|)$ time complexity using the power iteration method.

\item \textbf{Eigenvector Centrality (EC).} An improvement over the simple degree centrality, EC takes into account both the degree of a node and the quality of the neighboring nodes (higher quality connection incurs higher centrality) \cite{eigenvector_centrality}. As the name suggests, computation of eigenvector centrality involves eigenvalue computations, and the centrality is formally defined as  $EC(v_i) = \frac{1}{\lambda_1} \sum_j A_{ij}x_j$, with $\lambda_1$ as the highest eigenvalue and $x_j$ as node $v_j$'s corresponding eigenvector. EC can be computed using the power iteration method with $O(|\mathcal{V}|+|\mathcal{E}|)$ \cite{CentralityTimeComplexity} time complexity.
\end{itemize}

\begin{figure}[t]
\centering
\includegraphics[width=1\columnwidth]{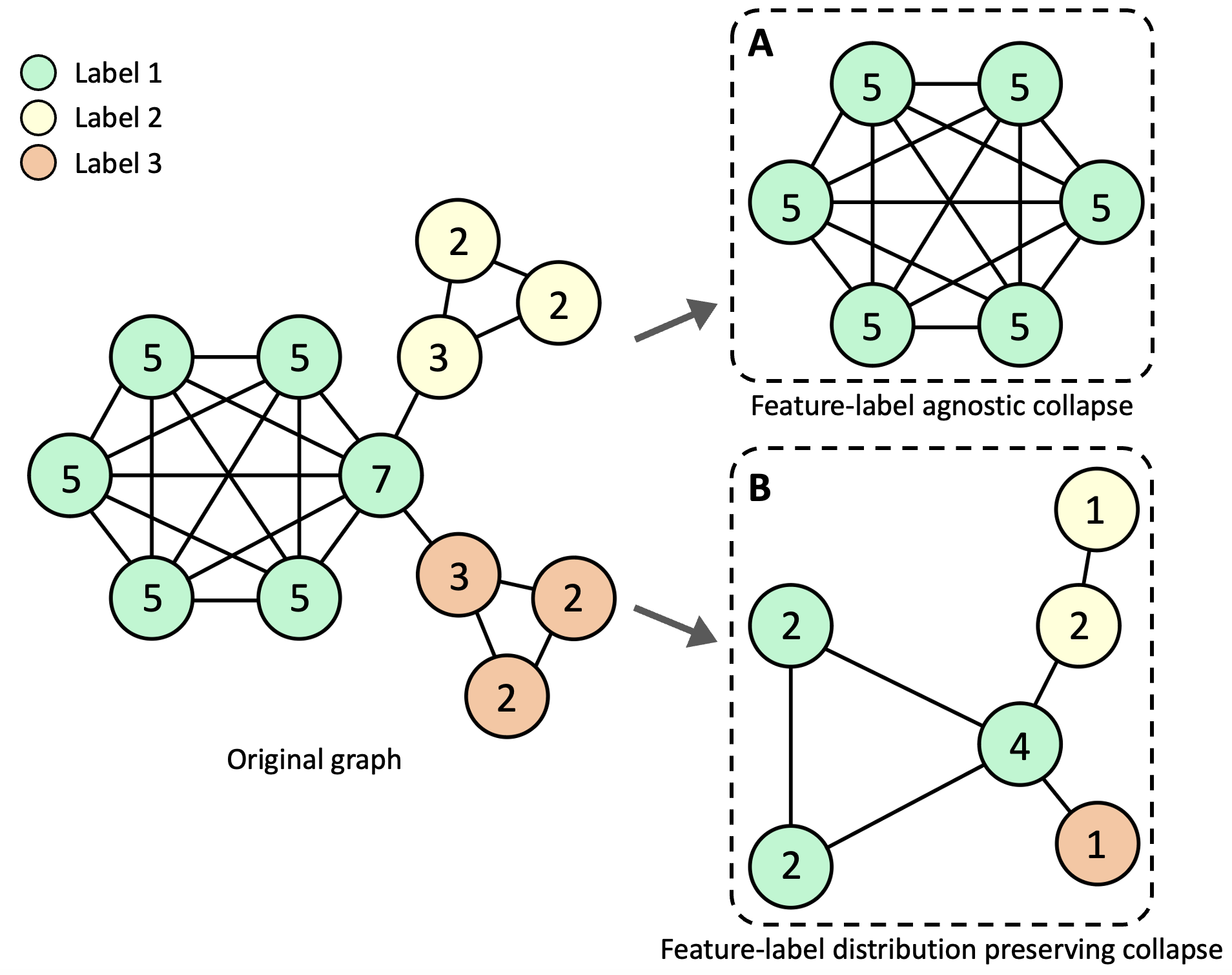}
\caption{Feature-label agnostic collapse (\textbf{A}) causes obliteration of labels with only inferior centrality measure nodes. Conversely, feature-label distribution preserving collapse (\textbf{B}) maintains the original label distribution, effectively avoiding complete label removal. The node color depicts the node label, while the node number depicts the centrality measure (degree centrality).}
\label{why_preserve}
\end{figure}

\textbf{Feature-Label Constraint.} While collapsing a graph purely based on centrality can theoretically maintain the graph's most important nodes, it remains agnostic to the distribution of node features and labels. Hence, there is no guarantee that the original feature and label distributions of the uncollapsed graph are preserved. In the worst case, we can have a certain set of training labels (or features) completely obliterated by the collapse if all of the nodes with the corresponding labels have weak centrality measures. The collapse depicted in annotation A of Figure~\ref{why_preserve} visualizes an example scenario where we have a set of labels entirely removed from the collapsed graph due to consisting only of weaker centrality nodes. Hence, to preserve the original distribution, we propose a clustering-based scheme that utilizes K-Means clustering and a modified Euclidean distance metric to cluster the nodes based on their similar features and labels. 

The main idea is to cluster the nodes such that each cluster represents a distinct group of nodes based on their features and labels. Afterward, we can \emph{distribute the ``collapse"} among these groups (e.g., suppose we are collapsing 50\% nodes and we have three feature-label clusters, we can collapse each cluster proportionally by $\sim$50\%). Provided we have enough clusters to capture the original feature-label distribution, the collapsed nodes' relative distribution will be similar to the original graph. The collapse depicted in annotation B of Figure~\ref{why_preserve} visualizes this feature-label distribution preserving collapse. 

\textbf{Dimension Normalized Euclidean Distance.} In the case of multi-label classification, the most straightforward way to cluster the nodes based on both similar features $X \in \mathbb{R}^{|\mathcal{V}| \times F}$ and labels $Y \in \mathbb{R}^{|\mathcal{V}| \times L}$ is by concatenating both and performing the clustering based on $[X, Y]$. Similarly, with multi-class classification, we can perform one hot encoding on the labels and concatenate it with the features. However, the concern with such approach is that the prioritization between features and labels is going to be affected by the dimensions of the features and labels (i.e., higher feature dimension will skew the Euclidean distance metric to prioritize features over labels, and vice versa, as shown in Equation~\ref{eqn:original_euclidean2}).
\begin{equation}\label{eqn:original_euclidean2}
    \resizebox{.82\hsize}{!}{
        $D(v_i,v_j)^2 = \sum_{f}^{F}{(x_{if} - x_{jf})^2} + \sum_{l}^{L}{(y_{il} - y_{jl})^2}$
    }
\end{equation}

Where $v_i$ denotes node $i$, $x_{if}$ denotes the $f$\textsuperscript{th} feature of node $i$ , and $y_{il}$ denotes the $l$\textsuperscript{th} label of node $i$.

Therefore, we further introduce a normalization scheme on the Euclidean distance with an added hyperparameter to denote the prioritization, $\gamma \in [0,1]$ as shown in Equation~\ref{eqn:new_euclidean} and \ref{eqn:new_euclidean2}:
\begin{equation}\label{eqn:new_euclidean}
    \resizebox{.88\hsize}{!}{
            $D(v_i,v_j)^2 = \alpha\sum_{f}^{F}{(x_{if}-x_{jf})^2} + \beta\sum_{l}^{L}{(y_{il}-y_{jl})^2}$
    }
\end{equation}
\begin{equation}\label{eqn:new_euclidean2}
\resizebox{.7\hsize}{!}{
        $\alpha = \gamma\frac{\mathrm{max}(D,L)}{D} \;\;\;\;\;\;
        \beta = (1-\gamma)\frac{\mathrm{max}(D,L)}{L}$
        }
\end{equation}

Here, $\gamma = 0$ means that we are only considering the labels, while $\gamma = 1$ means only features are being considered. We can use $\gamma = 0.5$ to consider both features and labels equally. Note that both features and labels need to be scale-normalized to fall between 0 to 1 inclusive.

We can rewrite Equation~\ref{eqn:new_euclidean} by including the $\alpha$ and $\beta$ terms as part of the features $x$ and labels $y$ as shown in Equation~\ref{eqn:new_euclidean3}. This means we can apply element-wise multiplication to the features and labels with $\sqrt\alpha$ and $\sqrt\beta$, respectively. Then we can use any clustering algorithm such as K-Means based on our proposed node-to-node distance. Formally, we define $M \in \mathbb{R}^{|\mathcal{V}| \times (F + L)}$ as the normalized feature label concatenation, which is computed using Equation~\ref{eqn:new_euclidean4}. 

\begin{equation}\label{eqn:new_euclidean3}
   \resizebox{.9\hsize}{!}{
            $D(v_i,v_j)^2 = \sum_{f}^{F}{(\sqrt{\alpha}x_{if}-\sqrt{\alpha}x_{jf})^2} + 
            \sum_{l}^{L}{(\sqrt{\beta}y_{il}-\sqrt{\beta}y_{jl})^2}$}
\end{equation}
\begin{equation}\label{eqn:new_euclidean4}
\resizebox{0.5\hsize}{!}{
        $M = (\sqrt\alpha \odot X)\oplus(\sqrt\beta \odot Y)$
        }
\end{equation}

Next, we cluster the nodes based on $M$ to get the relatively distinct node groups based on their features and labels which we can use to balance the cluster proportions. We then collapse each cluster using our topology-aware collapse method as illustrated in Figure~\ref{main_figure}. The resulting collapsed graph naturally maintains the original feature and label distribution. FALCON algorithm is given in appendix A.

\textbf{FALCON Algorithm.} Finally, we present the pseudocode for the proposed FALCON (Algorithm~\ref{alg:falcon_algo}). The parameters $\Psi$, $\eta$, $\gamma$, and $\zeta$ denote the training node budget, number of feature-label clusters, feature-label prioritization constant, and centrality metric used, respectively.

\begin{algorithm}[h!]
\caption{FALCON}
\label{alg:falcon_algo}
\textbf{Input}: $\mathcal{G}=\{\mathcal{V},\mathcal{E},X,Y\}$\\
\textbf{Parameter}: $\Psi,\eta,\gamma,\zeta$\\
\textbf{Output}: $\mathcal{G}'=\{\mathcal{V}',\mathcal{E}',X',Y'\}$
\begin{algorithmic}[1] 
\STATE Compute centrality measure $\Phi$ of the original graph $\{\mathcal{V},\mathcal{E}\}$ using $\zeta$
\STATE Compute $\alpha$ and $\beta$ using Equation~\ref{eqn:new_euclidean2}
\STATE Compute $M$ using Equation~\ref{eqn:new_euclidean4}
\STATE Obtain the feature-label node clusters [$\mathcal{V}_{1}, \mathcal{V}_{2}, ..., \mathcal{V}_{\eta}$] using K-Means clustering with $M$ as the feature
\STATE Obtain cluster node budgets [$\Psi_{1}, \Psi_{2}, ..., \Psi_{\eta}$] by distributing $\Psi$ proportionally to each feature-label node clusters
\FOR{i $\leftarrow$ 1 to $\eta$}
\STATE Sort nodes in $\mathcal{V}_{i}$ based on $\Phi$, in an ascending order
\WHILE {$|\mathcal{V}_{i}| > \Phi_{i}$}
\STATE Get node $v_k$ with the lowest $\Phi$ from $\mathcal{V}_{i}$
\STATE Find $v_s$ such that $v_s$ is a neighbouring node of $v_k$ with the highest $\Phi$ (note that $v_s$ can be outside of $\mathcal{V}_{i}$)
\STATE Merge $v_k$ to $v_s$ by moving all edges connected to $v_k$ to $v_s$ and removing $v_k$ from $\mathcal{G}$
\ENDWHILE
\ENDFOR
\STATE \textbf{return} $\mathcal{G}$
\end{algorithmic}
\end{algorithm}

\section {Memory-Efficient GNN Framework}

This section details the choice of the mini-batch construction and layer compression stages in the 4-stage memory-efficient GNN framework (Figure~\ref{training_stages}). This framework is used to further minimize GPU memory footprint as much as possible in addition to FALCON's graph reduction and to demonstrate the holistic potential of combining FALCON with other SOTA memory-efficient GNN methods. As shown in Figure~\ref{main_figure}, the memory-efficient framework comprises FALCON for the graph reduction, SIGN model for the mini-batching construction, and activation quantization for the layer-wise compression. However, application of FALCON is not limited only to SIGN model as shown in our experiments later on where we apply FALCON to other SOTA GNN models.

\textbf{Mini-batch construction stage.} Here, many recent SOTA works focus on mini-batching in coupled feature aggregation and transformation models (e.g., GCN). However, most of these methods have limitations, such as the removal of out-of-batch edges (e.g., Cluster-GCN \cite{cluster-gcn}) or involve expensive CPU to GPU data transfer for out-of-batch edges (e.g., GNN AutoScale \cite{cluster-gcn} and LMC-GNN \cite{LMC-GNN}). The roots of these limitations lie in the coupling of feature aggregation and transformation phases. Taking a different direction, SIGN \cite{frasca2020sign} and SGC \cite{SGC} decouple the feature aggregation from the feature transformation phase under the assumption that non-linearity between layers is not essential in GCNs. SGC pioneers this idea, and SIGN extend it by providing a general decoupled GNN model through the concatenation of various aggregated features and the use of general MLP to replace the single-layer feature transformation phase in SGC. A simplified, yet more expressive version of SIGN is introduced by \cite{frasca2020sign} and is shown in Equation~\ref{eqn:newSIGN_main} and \ref{eqn:newSIGN_main2} (the original SIGN implementation is given in appendix B):
\begin{equation}\label{eqn:newSIGN_main}
\resizebox{.3\hsize}{!}{
    $\hat{Y} = \mathbf{MLP}(Z, \theta)$
    }
\end{equation}
\begin{equation}\label{eqn:newSIGN_main2}
    Z = [IX^{0}, \tilde{A}X^{0}, ..., \tilde{A}^{n}X^{0}]
\end{equation}

Where $Z$ is a concatenation of various $k$-hop aggregated features (multiplication of power matrices $\tilde{A}^{k}$ and $X^{0}$).

Since the feature transformation phase uses independent data points, mini-batching can be done trivially, which circumvents the aforementioned limitations. The generalization to multiple aggregations and the use of MLP for the transformation phase also puts it ahead of SGC. On that account, we further incorporate SIGN in our framework for the mini-batch construction.

\textbf{Layer Compression Stage.} Moving to the layer compression stage, we quantize the data stored in each layer, further minimizing the GPU memory footprint. Most layers store both weight and activation, usually done in full 32-bit precision. Activation, however, tends to be the major memory usage contributor as it depends on the number of samples. Consequently, ActNN \cite{chen2021actnn} pioneers the use of activation quantization on convolutional neural network (CNN) which is then introduced to several GNN methods (e.g., GCN, GAT, and Cluster-GCN) by EXACT \cite{liu2022exact}. 
Here, we extend our framework using the 2-bit activation quantization by ActNN on the feature transformation MLP layers, and we use the term QSIGN to denote the activation quantized SIGN model. 

The activation quantization by \cite{chen2021actnn} is defined as $quant(\mathbf{h_{ig}})=\frac{\mathbf{h_{ig}}-Z_{ig}}{R_{ig}}.B$, where $h_{ig}$ denotes the embeddings of node $v_i$ belonging to group $g$ (if quantization is done on multiple groups of embeddings), $Z_{ig}=$min$(\mathbf{h_{ig}})$, $R_{ig}=$max$(\mathbf{h_{ig}})-$min$(\mathbf{h_{ig}})$, and $B=2^b-1$, with $b$ denoting the number of compression bits which in our case is 2. Dequantization is the inverse of the quantization and is defined as $dequant(\mathbf{h_{ig}})=Z_{ig}+quant(\mathbf{h_{ig}}).\frac{R_{ig}}{B}$.

\section{Experiments}

\textbf{Evaluation Datasets.} We evaluate FALCON on four sufficiently large public datasets (PPI \cite{PPI}, Flickr \cite{RefWorks:RefID:14-zeng2019graphsaint:,graphsaint}, MedMNIST Organ-S \cite{medmnistv2}, and MedMNIST Organ-C \cite{medmnistv2}) to highlight the GPU memory impact. Organ-S and Organ-C datasets were originally collected for an image classification task, with each sample having 28 by 28 single-channel pixels. To represent it as a graph, we use the vectorized pixel intensities as the node feature and construct the edges based on the features' cosine similarity (to ensure the sparsity of the graph, we set the similarity threshold at $\sim$1.2 million edges). The dataset properties for the benchmark are shown in Table~\ref{tab:dataset_details} along with the hyperparameters. Here, $
\gamma$ and number of hops (the number of aggregation hops for SIGN) are determined using grid search based on validation accuracy with a range of 0.4 to 0.6 and 1 to 6, respectively. On the other hand, we set the number of layers and hidden channels sufficiently large to mimic complex model training and to highlight the GPU memory impact of the training. Note that all datasets are trained inductively, and all results are based on a 95\% confidence interval of 5 runs. All experiments are done on a Linux machine with a Ryzen 5 5600H CPU, 16 GB DDR4 memory, and a 4GB NVIDIA RTX 3050ti GPU.

\begin{table}[htbp]
  \centering
  \caption{Dataset details and hyperparameters.}
  \scalebox{0.8}{
    \begin{tabular}{l|c|c|c|c}
          & PPI   & \multicolumn{1}{p{4.665em}|}{MedMNIST Organ-S} & \multicolumn{1}{p{4.665em}|}{MedMNIST Organ-C} & Flickr \\
    \midrule
    \# of Nodes & 56944 & 25221 & 23660 & 89250 \\
    \# of Edges & 1587264 & 1276046 & 1241622 & 899756 \\
    \# of Features & 50    & 784   & 784   & 500 \\
    \# of Labels & 121   & 11    & 11    & 7 \\
    Task Type & Multi-label & Multi-class & Multi-class & Multi-class \\
    \midrule
    Training Nodes & 44906 & 13940 & 13000 & 44625 \\
    Validation Nodes & 6514  & 2452  & 2392  & 22312 \\
    Test Nodes & 5524  & 8829  & 8268  & 22313 \\
    \midrule
    \# of layers & 3     & 3     & 3     & 3 \\
    Hidden channels & 1024  & 2048  & 2048  & 2048 \\
    Dropouts & 0.2   & 0.5   & 0.5   & 0.5 \\
    \# of Hops (SIGN) & 2     & 1     & 1     & 5 \\
    \# of Mini-batch & 10    & 3     & 3     & 5 \\
    $\gamma$ & 0.52  & 0.54  & 0.49  & 0.49 \\
    \# of FALCON Clusters & 100   & 100   & 100   & 100 \\
    Learning Rate  & 0.005 & 0.0005 & 0.0005 & 0.0005 \\
    \end{tabular}%
    }
  \label{tab:dataset_details}%
\end{table}%

\textbf{Benchmark Methods.} We train various SOTA models (e.g., GCN \cite{GCN}, Cluster-GCN \cite{cluster-gcn}, GNN AutoScale \cite{GNNAutoScale:}, SIGN \cite{frasca2020sign}) along with their FALCON collapsed variant on several public datasets. For each FALCON variant, we also trained multiple models with different centrality type (e.g., degree, eigenvector, betweenness, closeness, PageRank). Although we are mainly interested in the GPU memory saving contributed by the FALCON collapse, we also measure the epoch time and various performance metric (i.e., for multi-label tasks, we use accuracy, micro F1, micro sensitivity, and micro specificity, while for multi-class tasks, we only use accuracy and micro specificity as the remaining two yields the same result as accuracy). Additionally, we also measure the impact of feature-label distribution preservation by comparing the distribution error and performance of FALCON against pure centrality-based collapse \cite{cqsign}.

\begin{table*}[htbp]
  \centering
  \caption{Prediction result comparison of FALCON and Pure Centrality-Based Collapse \cite{cqsign}.}
  \scalebox{0.9}{
    \begin{tabular}{p{10em}|>{\centering\arraybackslash}p{4.2em}>{\centering\arraybackslash}p{4.2em}>{\centering\arraybackslash}p{4.2em}>{\centering\arraybackslash}p{4.2em}|>{\centering\arraybackslash}p{4.2em}>{\centering\arraybackslash}p{4.2em}|>{\centering\arraybackslash}p{4.2em}>{\centering\arraybackslash}p{4.2em}|>{\centering\arraybackslash}p{4.2em}>{\centering\arraybackslash}p{4.2em}}
    \multirow{2}[1]{*}{Methods} & \multicolumn{4}{c|}{PPI}      & \multicolumn{2}{c|}{MedMNIST Organ-S} & \multicolumn{2}{c|}{MedMNIST Organ-C} & \multicolumn{2}{c}{Flickr} \\
    \multicolumn{1}{c|}{} & ACC   & F1    & Sensitivity & Specificity & ACC   & Specificity & ACC   & Specificity & ACC   & Specificity \\
    \midrule
    FALCON-QSIGN (DC) & \textbf{80.73±0.10} & 61.54±0.32 & 51.50±0.50 & \textbf{93.23±0.16} & \textbf{60.75±0.25} & \textbf{96.08±0.03} & \textbf{77.49±0.35} & \textbf{97.75±0.04} & \textbf{49.13±0.31} & \textbf{91.52±0.05} \\
    FALCON-QSIGN (BC) & \textbf{87.67±0.06} & 77.37±0.11 & 70.44±0.19 & \textbf{95.02±0.11} & \textbf{60.66±0.62} & \textbf{96.07±0.06} & 77.72±0.58 & 97.77±0.06 & \textbf{48.74±0.31} & \textbf{91.46±0.05} \\
    FALCON-QSIGN (PR) & \textbf{80.30±0.05} & 60.24±0.09 & 49.85±0.16 & \textbf{93.31±0.11} & \textbf{60.87±0.25} & \textbf{96.09±0.03} & \textbf{77.70±0.28} & \textbf{97.77±0.03} & 48.44±0.38 & 91.41±0.06 \\
    FALCON-QSIGN (CC) & \textbf{97.33±0.03} & \textbf{95.46±0.04} & 93.70±0.09 & \textbf{98.89±0.06} & \textbf{60.41±0.30} & \textbf{96.04±0.03} & \textbf{77.82±0.51} & \textbf{97.78±0.05} & -     & - \\
    FALCON-QSIGN (EC) & \textbf{98.64±0.03} & \textbf{97.71±0.05} & \textbf{96.90±0.10} & 99.38±0.07 & \textbf{60.55±0.44} & \textbf{96.05±0.04} & \textbf{78.02±0.33} & \textbf{97.80±0.03} & 48.05±0.35 & 91.34±0.06 \\
    \midrule
    \midrule
    C-QSIGN (DC) \cite{cqsign} & 80.26±0.09 & \textbf{66.52±0.22} & \textbf{65.51±0.39} & 86.56±0.10 & 59.93±0.49 & 95.99±0.05 & 77.37±0.26 & 97.74±0.03 & 48.89±0.37 & 91.48±0.06 \\
    C-QSIGN (BC) \cite{cqsign} & 87.29±0.05 & \textbf{77.59±0.09} & \textbf{73.49±0.35} & 93.19±0.18 & 60.01±0.45 & 96.00±0.04 & \textbf{77.80±0.56} & \textbf{97.78±0.06} & 48.71±0.17 & 91.45±0.03 \\
    C-QSIGN (PR) \cite{cqsign} & 79.78±0.11 & \textbf{64.41±0.19} & \textbf{61.13±0.72} & 87.75±0.44 & 59.93±0.49 & 95.99±0.05 & 77.37±0.26 & 97.74±0.03 & \textbf{48.66±0.41} & \textbf{91.44±0.07} \\
    C-QSIGN (CC) \cite{cqsign} & 97.13±0.02 & 95.13±0.03 & \textbf{93.70±0.06} & 98.59±0.04 & 59.93±0.49 & 95.99±0.05 & 77.37±0.26 & 97.74±0.03 & -     & - \\
    C-QSIGN (EC) \cite{cqsign} & 98.63±0.03 & 97.68±0.05 & 96.83±0.06 & \textbf{99.39±0.03} & 59.93±0.49 & 95.99±0.05 & 77.37±0.26 & 97.74±0.03 & \textbf{48.28±0.37} & \textbf{91.38±0.06} \\
    \end{tabular}%
    }
  \label{tab:benchmark_falco_vs_agnostic}%

\begin{minipage}{2\columnwidth}
\vspace{0.1cm}
\vspace{0.1cm}
\small  Notes: Results are \textbf{bolded} if they are better than their counterparts. Further comparisons are included in appendix E.
\end{minipage}
\end{table*}%

\begin{figure}[t]
\centering
\includegraphics[width=1.0\columnwidth]{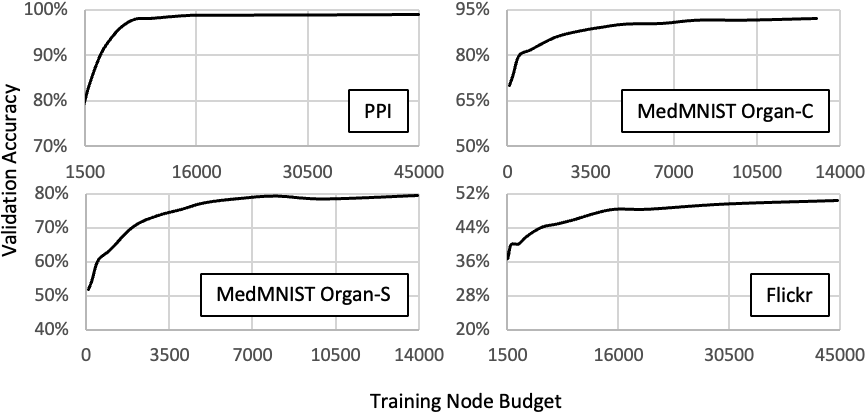}
\caption{Validation accuracy convergence on training node budgets by FALCON-CQSIGN (EC).}
\label{convergence}
\end{figure}

\textbf{Node Budget Convergence.} First, we evaluate the impact of FALCON while varying training node budgets on its validation accuracy. We train FALCON-QSIGN (EC) models with varying training node budgets and plot the validation accuracy as depicted in Figure~\ref{convergence}. Our experiment shows that convergence is achieved with approximately 15000 training nodes for PPI and Flickr and 8000 training nodes for Organ-C and Organ-S. Being collapsible to as low as $\sim$34\%  (PPI and Flickr) and $\sim$58\% (Organ-C and Organ-S) of the original graph while maintaining their predictive performance implies that these datasets are highly collapsible.

\textbf{Prediction Quality Comparison.} Next, we benchmark both the training resource usage and prediction results on test set of FALCON on various GNN models. Table~\ref{tab:main_result} shows the prediction quality of our benchmark and suggests that across all benchmark models, FALCON can achieve very similar performance to their uncollapsed variant while using only fractions of the complete training graph. The results also suggest the choice of centrality measure can significantly impact prediction quality on some datasets (e.g., PPI). This implies that the choice of centrality needs to be treated as a hyperparameter. Nonetheless, our benchmark suggests that EC performs the best overall as it performs very closely to a complete graph and is also scalable to larger graphs due to its linear time complexity \cite{CentralityTimeComplexity}. Additionally, our framework (FALCON-QSIGN) produces similar prediction results to other benchmark models.

\textbf{Computational Resource Comparison.} The main benefit of our proposed FALCON and FALCON-QSIGN framework is the training computational resource efficiency evidenced in Table~\ref{tab:main_resource}. This result shows that FALCON can significantly reduce GPU memory usage and epoch time during training across all models. Furthermore, FALCON-QSIGN framework has the lowest GPU memory usage and 2\textsuperscript{nd} lowest epoch time (excluding GCN) across all benchmarks. Being outperformed by SIGN in epoch time shows the quantization overhead required to encode and decode the activation.

\textbf{Feature-Label Constraint Impact.} Here, we show the importance of preserving feature and label distribution in the post-collapsed graph instead of doing pure centrality-based collapse as in \cite{cqsign}. The post-collapse macro average label distribution error can be computed using $L_{err} = \frac{1}{L}\sum_{l}^{L}(|\frac{n_l}{n}-\frac{N_l}{N}|)$, where $\frac{n_l}{n}$ denotes the node ratio with label $l$ in the collapsed graph, while $\frac{N_l}{N}$ denotes the same label ratio, but for the original graph. Table~\ref{tab:benchmark_label_error} shows this label distribution error, which suggests that FALCON preserves label distribution best across all benchmarks. Furthermore, while label distribution error is not the main evaluation metric we use, this result is also reflected in the graph prediction results as suggested by Table~\ref{tab:benchmark_falco_vs_agnostic}, which shows that FALCON has better overall prediction performance compared to its feature label agnostic counterparts. 


\textbf{Graph Coarsening Comparison.} Finally, we benchmark FALCON against the existing SOTA graph reduction method using coarsening algorithm \cite{huang2021scaling}. However, as we encountered memory errors when coarsening our primary datasets using the available source code, we primarily use the publicly available Cora and PubMed datasets \cite{yang2016revisiting} as done by \cite{huang2021scaling}. Here, the hyperparameter is set in a similar way to the main benchmark, and we use a 3-layer network with 1536 hidden layers, $\gamma=0.5$, 100 FALCON clusters, dropouts of 0.5, and a learning rate of 0.0005 for both Cora and PubMed dataset. Table~\ref{tab:benchmark_falco_vs_coarsen} depicts the prediction results benchmark, which shows that FALCON with the optimal centrality measure (i.e., PR in this case) is able to outperform the graph coarsening method. Moreover, we also compare the label distribution error of the post-collapsed graph in Table~\ref{tab:benchmark_label_error}, which shows that our method significantly outperforms \cite{huang2021scaling} in preserving the original feature and label distribution. Initially, we also intended to benchmark against the graph reduction method by \cite{scalablegraphreduction}. However, their source code is not publicly available at the time of writing.

\section{Conclusion}

We present FALCON, a topology-aware graph-collapsing scheme that preserves the original graph feature and label distributions for memory-efficient GNN training. Our benchmark shows that FALCON significantly reduces the training GPU memory footprint (up to 66\% reduction on PPI) and epoch time (up to 72\% reduction on Flickr) across different GNN models and four public datasets while maintaining similar performance to the original graph. Additionally, we introduce a memory-efficient GNN framework that combines various SOTA methods (FALCON-QSIGN). Experimental result showed that FALCON-QSIGN has the lowest GPU memory footprint compared to other benchmarked methods with a phenomenal 97\% reduction compared to GCN on the PPI dataset, and a lowest reduction of 43\% compared to SIGN on the MedMNIST Organ-C dataset.

\begin{table}[htbp]
  \centering
  \caption{Prediction result comparison of FALCON and coarsening \cite{huang2021scaling}.}
  \scalebox{0.8}{
    \begin{tabular}{@{}p{12em}|>{\centering\arraybackslash}p{4.2em}>{\centering\arraybackslash}p{4.2em}|>{\centering\arraybackslash}p{4.2em}>{\centering\arraybackslash}p{4.2em}@{}}
    \multirow{2}[0]{*}{Methods} & \multicolumn{2}{c|}{Cora} & \multicolumn{2}{c}{PubMed} \\
    \multicolumn{1}{c|}{} & ACC   & Specificity & ACC   & Specificity \\
    \midrule
    QSIGN & \underline{\textbf{87.82±0.49}} & \underline{\textbf{97.97±0.08}} & \underline{\textbf{90.12±0.24}} & \underline{\textbf{95.06±0.12}} \\
    FALCON-QSIGN (DC) & \underline{84.16±0.84} & \underline{97.36±0.14} & 87.52±0.52 & 93.76±0.26 \\
    FALCON-QSIGN (BC) & 83.60±0.82 & 97.27±0.14 & 85.92±0.80 & 92.96±0.40 \\
    FALCON-QSIGN (PR) & \textbf{84.36±0.62} & \textbf{97.39±0.10} & \textbf{88.24±0.24} & \textbf{94.12±0.12} \\
    FALCON-QSIGN (CC) & 82.46±0.75 & 97.08±0.13 & 86.56±1.50 & 93.28±0.75 \\
    FALCON-QSIGN (EC) & 81.92±0.38 & 96.99±0.06 & 87.18±0.23 & 93.59±0.11 \\
    Coarsen-QSIGN & 81.82±0.53 & 96.97±0.09 & \underline{87.54±0.30} & \underline{93.77±0.15} \\
    \midrule
    GCN \cite{GCN}   & \underline{\textbf{86.52±0.21}} & \underline{\textbf{97.75±0.03}} & \underline{\textbf{87.52±0.13}} & \underline{\textbf{93.76±0.07}} \\
    FALCON-GCN (DC) & \underline{83.62±0.59} & \underline{97.27±0.10} & 84.20±0.11 & 92.10±0.05 \\
    FALCON-GCN (BC) & 82.20±0.96 & 97.03±0.16 & 85.16±0.21 & 92.58±0.11 \\
    FALCON-GCN (PR) & \textbf{84.50±0.67} & \textbf{97.42±0.11} & \underline{84.44±0.12} & \underline{92.22±0.06} \\
    FALCON-GCN (CC) & 82.66±0.78 & 97.11±0.13 & 82.08±0.42 & 91.04±0.21 \\
    FALCON-GCN (EC) & 81.96±1.16 & 96.99±0.19 & 82.90±0.12 & 91.45±0.06 \\
    Coarsen-GCN \cite{huang2021scaling} & 81.74±0.65 & 96.96±0.11 & \textbf{85.36±0.10} & \textbf{92.68±0.05} \\
    \end{tabular}%
    }
  \label{tab:benchmark_falco_vs_coarsen}%
\begin{minipage}{1\columnwidth}
\vspace{0.1cm}
\vspace{0.1cm}
\small  Notes: Cora is collapsed to $\sim$41\% and Pubmed to $\sim$27\%. \underline{\textbf{Bold underline}}, \textbf{bold}, and  \underline{underline} denotes 1\textsuperscript{st}, 2\textsuperscript{nd}, and 3\textsuperscript{rd} best result of each models, respectively.
\end{minipage}
\end{table}%

\begin{table}[htbp]
  \centering
  \caption{Post-collapse label distribution error comparison of various graph reduction methods.}
  \scalebox{0.85}{
    \begin{tabular}{@{}p{7.1em}|>{\centering\arraybackslash}p{2.5em}>{\centering\arraybackslash}p{3.5em}>{\centering\arraybackslash}p{3.5em}>{\centering\arraybackslash}p{2.5em}>{\centering\arraybackslash}p{3em}>{\centering\arraybackslash}p{3em}@{}}
    \multicolumn{1}{c|}{\multirow{2}[1]{*}{Methods}} & \multicolumn{6}{c}{Macro Averaged Label Distribution Error} \\
          & PPI   & Organ-S & Organ-C & Flickr & Cora  & PubMed \\
    \midrule
    FALCON (EC) & \textbf{0.98\%} & \textbf{0.14\%} & \textbf{0.16\%} & \textbf{0.06\%} & \textbf{3.62\%} & \textbf{0.15\%} \\
    Collapse (EC) & 2.74\% & 18.65\% & 18.75\% & 17.82\% & 6.80\% & 5.63\% \\
    Coarsen \cite{huang2021scaling} & -     & -     & -     & -     & 25.68\% & 3.15\% \\
    \end{tabular}%
    }
  \label{tab:benchmark_label_error}%
\begin{minipage}{1\columnwidth}
\vspace{0.1cm}
\vspace{0.1cm}
\small  Notes: Lowest is \textbf{bolded}. Here, collapse denotes feature-label agnostic collapse.
\end{minipage}
\end{table}%

\begin{table*}[htbp]
  \centering
  \caption{Prediction results by FALCON and benchmarks on various models and datasets.}
  
  
  \scalebox{0.8}{
    \begin{tabular}{p{13.6em}|>{\centering\arraybackslash}p{4.2em}>{\centering\arraybackslash}p{4.2em}>{\centering\arraybackslash}p{4.2em}>{\centering\arraybackslash}p{4.2em}|>{\centering\arraybackslash}p{4.2em}>{\centering\arraybackslash}p{4.2em}|>{\centering\arraybackslash}p{4.2em}>{\centering\arraybackslash}p{4.2em}|>{\centering\arraybackslash}p{4.2em}>{\centering\arraybackslash}p{4.2em}}
    \multirow{2}[1]{*}{Methods} & \multicolumn{4}{c|}{PPI}      & \multicolumn{2}{c|}{MedMNIST Organ-S} & \multicolumn{2}{c|}{MedMNIST Organ-C} & \multicolumn{2}{c}{Flickr} \\
    \multicolumn{1}{c|}{} & ACC   & F1    & Sensitivity & Specificity & ACC   & Specificity & ACC   & Specificity & ACC   & Specificity \\
    \midrule
    GCN \cite{GCN}   & \underline{\textbf{99.44±0.01}} & \underline{\textbf{99.06±0.01}} & \underline{\textbf{98.84±0.03}} & \underline{\textbf{99.70±0.01}} & \underline{\textbf{60.12±0.08}} & \underline{\textbf{96.01±0.01}} & \underline{\textbf{77.68±0.35}} & \underline{\textbf{97.77±0.04}} & OOM   & OOM \\
    FALCON-GCN (DC) & 82.17±0.91 & 68.67±1.72 & 65.30±1.87 & 89.38±0.51 & 58.80±0.56 & 95.88±0.06 & 74.61±0.48 & 97.46±0.05 & \underline{\textbf{48.07±0.54}} & \underline{\textbf{91.34±0.09}} \\
    FALCON-GCN (BC) & 88.98±0.33 & 80.37±0.69 & 75.38±1.04 & 94.79±0.07 & \underline{59.04±0.90} & \underline{95.90±0.09} & \underline{75.15±0.51} & \underline{97.52±0.05} & \underline{47.66±0.56} & \underline{91.28±0.09} \\
    FALCON-GCN (PR) & 81.93±0.52 & 66.40±1.45 & 59.68±2.13 & 91.44±0.34 & 58.99±0.41 & 95.90±0.04 & 74.76±0.34 & 97.48±0.03 & \textbf{47.95±0.54} & \textbf{91.33±0.09} \\
    FALCON-GCN (CC) & \underline{97.69±0.12} & \underline{96.10±0.20} & \underline{94.93±0.29} & \underline{98.87±0.08} & 58.99±0.43 & 95.90±0.04 & 74.57±0.62 & 97.46±0.06 & -     & - \\
    FALCON-GCN (EC) & \textbf{98.76±0.02} & \textbf{97.91±0.03} & \textbf{97.33±0.10} & \textbf{99.36±0.04} & \textbf{59.06±0.56} & \textbf{95.91±0.06} & \textbf{75.54±0.27} & \textbf{97.55±0.03} & 47.34±0.80 & 91.22±0.13 \\
    \midrule
    GAS \cite{GNNAutoScale:}   & \underline{\textbf{99.53±0.01}} & \underline{\textbf{99.22±0.02}} & \underline{\textbf{99.05±0.03}} & \underline{\textbf{99.74±0.02}} & \underline{58.19±0.29} & \textbf{95.82±0.03} & \underline{\textbf{76.03±1.16}} & \underline{\textbf{97.60±0.12}} & OOM   & OOM \\
    FALCON-GAS (DC) & 78.21±0.70 & 60.02±0.81 & 54.61±0.45 & 88.30±1.01 & 57.36±0.78 & 95.74±0.08 & 74.23±1.15 & 97.42±0.11 & 47.24±0.87 & 91.21±0.14 \\
    FALCON-GAS (BC) & 87.61±0.15 & 77.46±0.30 & 71.08±0.49 & 94.68±0.20 & 57.89±0.25 & 95.79±0.02 & 74.15±0.85 & 97.41±0.09 & \underline{47.43±0.79} & \underline{91.24±0.13} \\
    FALCON-GAS (PR) & 79.07±0.20 & 57.65±0.73 & 47.62±1.35 & 92.51±0.63 & 58.02±0.71 & 95.80±0.07 & \underline{74.33±0.79} & \underline{97.43±0.08} & \underline{\textbf{47.99±0.87}} & \underline{\textbf{91.33±0.15}} \\
    FALCON-GAS (CC) & \underline{96.80±0.12} & \underline{94.56±0.20} & \underline{92.94±0.16} & \underline{98.44±0.10} & \underline{\textbf{58.57±0.19}} & \underline{\textbf{95.86±0.02}} & \textbf{74.88±0.44} & \textbf{97.49±0.04} & -     & - \\
    FALCON-GAS (EC) & \textbf{98.41±0.10} & \textbf{97.32±0.17} & \textbf{96.79±0.27} & \textbf{99.10±0.09} & \textbf{58.20±0.63} & \underline{95.82±0.06} & 73.97±0.85 & 97.40±0.08 & \textbf{47.51±0.35} & \textbf{91.25±0.06} \\
    \midrule
    ClusterGCN \cite{cluster-gcn} & \underline{\textbf{99.40±0.01}} & \underline{\textbf{98.99±0.01}} & \underline{\textbf{98.79±0.02}} & \underline{\textbf{99.66±0.01}} & \textbf{60.11±0.41} & \textbf{96.01±0.04} & \underline{\textbf{78.09±1.37}} & \underline{\textbf{97.81±0.14}} & \underline{\textbf{46.42±0.14}} & \underline{\textbf{91.07±0.02}} \\
    FALCON-ClusterGCN (DC) & 84.66±0.08 & 71.60±0.15 & 64.61±0.29 & 93.23±0.14 & 59.53±0.75 & 95.95±0.08 & 76.10±1.65 & 97.61±0.16 & \underline{46.39±0.14} & \underline{\textbf{91.07±0.02}} \\
    FALCON-ClusterGCN (BC) & 90.12±0.14 & 82.64±0.27 & 78.55±0.38 & 95.06±0.07 & \underline{\textbf{60.43±0.87}} & \underline{\textbf{96.04±0.09}} & \textbf{77.16±1.63} & \textbf{97.72±0.16} & 46.33±0.12 & 91.05±0.02 \\
    FALCON-ClusterGCN (PR) & 84.01±0.09 & 70.25±0.21 & 63.06±0.40 & 92.96±0.16 & 59.25±0.94 & 95.92±0.09 & \underline{76.65±1.20} & \underline{97.67±0.12} & \textbf{46.40±0.15} & \underline{\textbf{91.07±0.02}} \\
    FALCON-ClusterGCN (CC) & \underline{96.65±0.28} & \underline{94.34±0.49} & \underline{93.21±0.61} & \underline{98.12±0.14} & \underline{59.59±0.66} & \underline{95.96±0.07} & 76.36±1.16 & 97.64±0.12 & -     & - \\
    FALCON-ClusterGCN (EC) & \textbf{97.95±0.40} & \textbf{96.56±0.67} & \textbf{96.14±0.71} & \textbf{98.72±0.27} & 59.23±0.94 & 95.92±0.09 & 76.15±1.14 & 97.62±0.11 & 46.12±0.21 & 91.02±0.04 \\
    \midrule
    SIGN \cite{frasca2020sign} & \underline{\textbf{99.45±0.01}} & \underline{\textbf{99.08±0.02}} & \underline{\textbf{98.83±0.02}} & \underline{\textbf{99.72±0.01}} & \underline{\textbf{61.37±0.47}} & \underline{\textbf{96.14±0.05}} & \underline{\textbf{79.21±0.23}} & \underline{\textbf{97.92±0.02}} & \underline{\textbf{50.54±0.35}} & \underline{\textbf{91.76±0.06}} \\
    FALCON-SIGN (DC) & 80.93±0.05 & 61.43±0.31 & 50.72±0.60 & 93.84±0.24 & \textbf{60.82±0.49} & \textbf{96.08±0.05} & \underline{77.95±0.47} & \textbf{97.80±0.05} & 48.43±0.62 & 91.40±0.10 \\
    FALCON-SIGN (BC) & 87.71±0.05 & 77.23±0.10 & 69.62±0.14 & 95.43±0.05 & 60.51±0.61 & 96.05±0.06 & \textbf{78.01±0.54} & \textbf{97.80±0.05} & \textbf{49.05±0.41} & \textbf{91.51±0.07} \\
    FALCON-SIGN (PR) & 80.31±0.06 & 59.16±0.17 & 47.63±0.34 & 94.27±0.18 & 60.59±0.40 & 96.06±0.04 & 77.93±0.37 & 97.79±0.04 & \underline{48.58±0.63} & \underline{91.43±0.10} \\
    FALCON-SIGN (CC) & \underline{97.39±0.02} & \underline{95.56±0.03} & \underline{93.66±0.07} & \underline{98.98±0.03} & \underline{60.68±0.18} & \underline{96.07±0.02} & 77.53±0.41 & 97.75±0.04 & -     & - \\
    FALCON-SIGN (EC) & \textbf{98.68±0.04} & \textbf{97.78±0.06} & \textbf{97.02±0.06} & \textbf{99.39±0.05} & 60.59±0.29 & 96.06±0.03 & 77.81±0.64 & 97.78±0.06 & 47.92±0.46 & 91.32±0.08 \\
    \midrule
    QSIGN & \underline{\textbf{99.42±0.01}} & \underline{\textbf{99.03±0.02}} & \underline{\textbf{98.76±0.02}} & \underline{\textbf{99.71±0.02}} & \underline{\textbf{61.80±0.57}} & \underline{\textbf{96.18±0.06}} & \underline{\textbf{79.25±0.32}} & \underline{\textbf{97.93±0.03}} & \underline{\textbf{50.53±0.16}} & \underline{\textbf{91.76±0.03}} \\
    FALCON-QSIGN (DC) & 80.73±0.10 & 61.54±0.32 & 51.50±0.50 & 93.23±0.16 & \underline{60.75±0.25} & \underline{96.08±0.03} & 77.49±0.35 & 97.75±0.04 & \textbf{49.13±0.31} & \textbf{91.52±0.05} \\
    FALCON-QSIGN (BC) & 87.67±0.06 & 77.37±0.11 & 70.44±0.19 & 95.02±0.11 & 60.66±0.62 & 96.07±0.06 & 77.72±0.58 & 97.77±0.06 & \underline{48.74±0.31} & \underline{91.46±0.05} \\
    FALCON-QSIGN (PR) & 80.30±0.05 & 60.24±0.09 & 49.85±0.16 & 93.31±0.11 & \textbf{60.87±0.25} & \textbf{96.09±0.03} & 77.70±0.28 & 97.77±0.03 & 48.44±0.38 & 91.41±0.06 \\
    FALCON-QSIGN (CC) & \underline{97.33±0.03} & \underline{95.46±0.04} & \underline{93.70±0.09} & \underline{98.89±0.06} & 60.41±0.30 & 96.04±0.03 & \underline{77.82±0.51} & \underline{97.78±0.05} & -     & - \\
    FALCON-QSIGN (EC) & \textbf{98.64±0.03} & \textbf{97.71±0.05} & \textbf{96.90±0.10} & \textbf{99.38±0.07} & 60.55±0.44 & 96.05±0.04 & \textbf{78.02±0.33} & \textbf{97.80±0.03} & 48.05±0.35 & 91.34±0.06 \\
    \end{tabular}%
    }
  \label{tab:main_result}%
\begin{minipage}{2\columnwidth}
\vspace{0.1cm}
\vspace{0.1cm}
\small  Notes: PPI \& Flickr are collapsed to $\sim$34\%, while Organ-C \& Organ-S to $\sim$58\%. GCN and GAS fail on Flickr due to an OOM error, while FALCON (CC) times out during the CC computation. \underline{\textbf{Bold underline}}, \textbf{bold}, and  \underline{underline} denotes 1\textsuperscript{st}, 2\textsuperscript{nd}, and 3\textsuperscript{rd} best result of each models, respectively.
\end{minipage}
\end{table*}%

\begin{table*}[htbp]
  \centering
  \caption{Computational resources of FALCON and benchmarks on various models and datasets.}
  \scalebox{0.8}{
    \begin{tabular}{p{13.6em}|>{\centering\arraybackslash}p{5.7em}>{\centering\arraybackslash}p{5em}|>{\centering\arraybackslash}p{5.7em}>{\centering\arraybackslash}p{5em}|>{\centering\arraybackslash}p{5.7em}>{\centering\arraybackslash}p{5em}|>{\centering\arraybackslash}p{5.7em}>{\centering\arraybackslash}p{5em}}
    \multirow{2}[0]{*}{Methods} & \multicolumn{2}{c|}{PPI} & \multicolumn{2}{c|}{MedMNIST Organ-S} & \multicolumn{2}{c|}{MedMNIST Organ-C} & \multicolumn{2}{c}{Flickr} \\
    \multicolumn{1}{c|}{} & GPU Memory & Epoch Time & GPU Memory & Epoch Time & GPU Memory & Epoch Time & GPU Memory & Epoch Time \\
    \midrule
    GCN \cite{GCN}   & 2347.3 & 3.0±1.9 & 1175.4 & 1.7±0.0 & 1102.4 & 1.9±0.1 & OOM   & OOM \\
    FALCON-GCN (DC) & 810.2 & 2.2±0.7 & 693.0 & 1.7±0.1 & 693.7 & 1.9±0.2 & 1221.2 & 1.9±0.1 \\
    FALCON-GCN (BC) & 808.0 & 2.2±0.7 & 691.1 & 1.7±0.1 & 693.0 & 1.9±0.1 & 1221.0 & 1.9±0.0 \\
    FALCON-GCN (PR) & 808.7 & 2.2±0.7 & 691.6 & 1.7±0.0 & 693.5 & 1.9±0.2 & 1221.1 & 1.9±0.0 \\
    FALCON-GCN (CC) & 807.6 & 2.1±0.7 & 692.9 & 1.8±0.2 & 693.3 & 1.8±0.2 & -     & - \\
    FALCON-GCN (EC) & 806.3 & 2.1±0.7 & 692.9 & 1.8±0.2 & 693.7 & 1.8±0.2 & 1220.7 & 1.9±0.0 \\
    \midrule
    GAS \cite{GNNAutoScale:}   & 427.3 & 737.3±4.2 & 545.5 & 329.3±1.9 & 519.6 & 308.2±1.0 & OOM   & OOM \\
    FALCON-GAS (DC) & 180.9 & 294.2±4.1 & 363.3 & 196.9±1.1 & 381.9 & 211.2±0.7 & 656.7 & 627.5±4.4 \\
    FALCON-GAS (BC) & 176.9 & 274.6±4.5 & 355.6 & 192.9±2.4 & 377.6 & 207.7±1.6 & 636.9 & 612.1±2.2 \\
    FALCON-GAS (PR) & 184.4 & 284.9±5.8 & 354.2 & 193.9±2.4 & 375.6 & 210.0±0.7 & 667.2 & 627.5±2.3 \\
    FALCON-GAS (CC) & 252.3 & 373.7±6.7 & 357.5 & 197.0±0.7 & 381.9 & 211.7±1.5 & -     & - \\
    FALCON-GAS (EC) & 239.3 & 389.6±2.8 & 360.5 & 196.9±1.4 & 382.4 & 211.2±2.4 & 582.7 & 514.0±3.8 \\
    \midrule
    ClusterGCN \cite{cluster-gcn} & 366.4 & 660.3±3.0 & 699.4 & 377.9±0.9 & 627.2 & 368.8±3.2 & 1133.2 & 817.0±2.5 \\
    FALCON-ClusterGCN (DC) & 149.9 & 274.3±0.8 & 437.4 & 229.5±2.6 & 438.8 & 262.7±2.9 & 457.2 & 291.2±1.8 \\
    FALCON-ClusterGCN (BC) & 154.0 & 258.5±1.2 & 432.7 & 233.9±10.1 & 433.9 & 255.3±2.8 & 452.6 & 289.8±1.5 \\
    FALCON-ClusterGCN (PR) & 152.5 & 277.8±1.9 & 446.1 & 240.4±6.6 & 436.3 & 260.9±2.8 & 463.3 & 290.3±0.7 \\
    FALCON-ClusterGCN (CC) & 149.0 & 227.0±3.0 & 432.3 & 229.6±2.3 & 434.4 & 261.7±3.1 & -     & - \\
    FALCON-ClusterGCN (EC) & 148.3 & 218.2±4.2 & 444.6 & 232.3±1.8 & 441.1 & 262.4±2.7 & 479.6 & 288.8±1.3 \\
    \midrule
    SIGN \cite{frasca2020sign}  & 177.3 & 176.1±0.9 & 402.3 & 173.7±1.7 & 382.9 & 164.6±1.6 & 763.7 & 1004.1±10.2 \\
    FALCON-SIGN (DC) & 81.4  & 64.9±0.1 & 275.6 & 99.6±0.9 & 275.7 & 99.2±0.6 & 350.9 & 288.6±3.8 \\
    FALCON-SIGN (BC) & 81.4  & 65.1±0.6 & 275.6 & 99.2±0.6 & 275.6 & 99.2±0.5 & 350.9 & 285.1±3.1 \\
    FALCON-SIGN (PR) & 81.4  & 65.2±0.6 & 275.6 & 99.3±0.6 & 275.7 & 99.4±0.6 & 350.9 & 285.1±2.6 \\
    FALCON-SIGN (CC) & 81.4  & 64.8±0.6 & 275.6 & 98.4±0.3 & 275.7 & 98.5±0.5 & -     & - \\
    FALCON-SIGN (EC) & 81.4  & 64.9±0.8 & 275.6 & 98.9±0.4 & 275.7 & 99.0±0.6 & 350.9 & 283.3±3.5 \\
    \midrule
    QSIGN & 116.9 & 273.8±0.5 & 278.4 & 229.1±1.0 & 269.4 & 218.4±1.0 & 566.5 & 1282.2±7.2 \\
    FALCON-QSIGN (DC) & \textbf{69.1} & 101.0±0.5 & \textbf{216.7} & 136.5±0.7 & \textbf{216.7} & 136.2±0.4 & \textbf{318.7} & 388.6±7.2 \\
    FALCON-QSIGN (BC) & \textbf{69.1} & 101.4±0.5 & \textbf{216.7} & 136.5±0.7 & \textbf{216.7} & 136.3±0.2 & \textbf{318.7} & 388.2±4.9 \\
    FALCON-QSIGN (PR) & \textbf{69.1} & 101.5±0.5 & \textbf{216.7} & 136.8±0.7 & \textbf{216.7} & 136.5±0.4 & \textbf{318.7} & 386.9±4.9 \\
    FALCON-QSIGN (CC) & \textbf{69.1} & 101.4±0.4 & \textbf{216.7} & 135.6±0.3 & \textbf{216.7} & 135.5±0.3 & -     & - \\
    FALCON-QSIGN (EC) & \textbf{69.1} & 101.4±0.6 & \textbf{216.7} & 136.0±0.5 & \textbf{216.7} & 136.1±0.2 & \textbf{318.7} & 384.4±3.7 \\
    \end{tabular}%
    }
  \label{tab:main_resource}%
\begin{minipage}{2\columnwidth}
\vspace{0.1cm}
\vspace{0.1cm}
\small  Notes: Epoch time is defined as the time for one complete pass of the dataset. Therefore, full-batched methods (i.e., GCN) will have a much shorter epoch time than mini-batched methods as no CPU to GPU data transfer is necessary in each epoch, and the gradient descent step is only done once per epoch. The GPU memory is in MB (overall lowest is \textbf{bolded}), and the epoch time is in milliseconds.
\end{minipage}
\end{table*}%

\clearpage
\begin{appendices}

\section{Original SIGN implementation}
The original implementation of SIGN adds a feature transformation layer for each of the aggregation as shown in Equation~\ref{eqn:SIGN_main} and \ref{eqn:SIGN_main2}:
\begin{equation}\label{eqn:SIGN_main}
\resizebox{.3\hsize}{!}{
    $\hat{Y} = \mathbf{MLP}(Z, \theta)$
    }
\end{equation}
\begin{equation}\label{eqn:SIGN_main2}
\resizebox{.6\hsize}{!}{
    $Z = \sigma([IX^{0}\theta_{0}, \tilde{A}X^{0}\theta_{1}, ..., \tilde{A}^{n}X^{0}\theta_{n}])$
    }
\end{equation}

Where $X$ denotes the original node features, $Z$ denotes the concatenated aggregated features, $\tilde{A}$ denotes the normalized adjacency matrix, $\theta$ denotes learnable weights, and $\hat{Y}$ denotes the prediction.


\section{Impact of Node Budgets to GPU Memory Usage and Epoch Time}

\begin{figure}[h!]
\centering
\includegraphics[width=1.0\columnwidth]{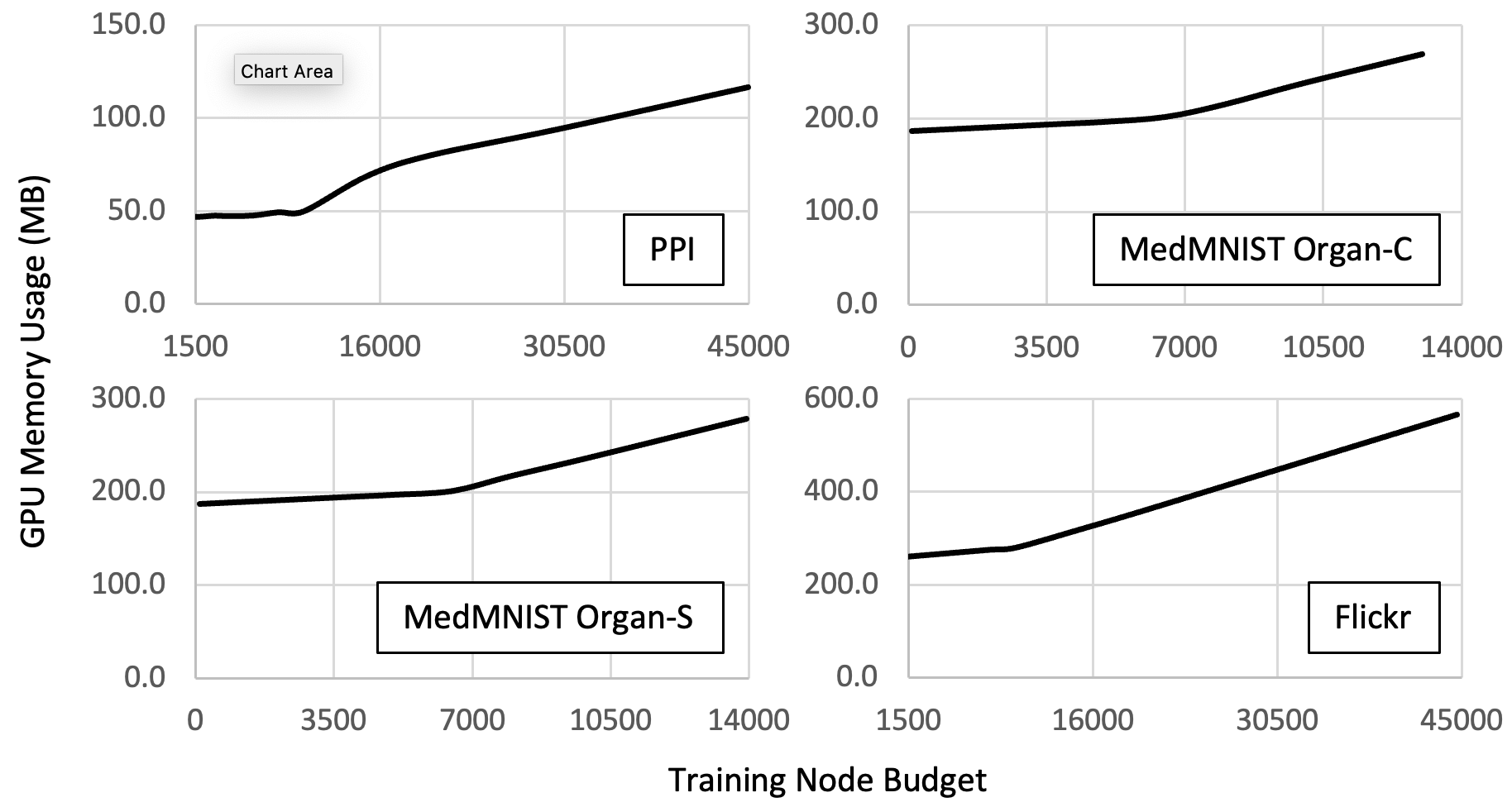}
\caption{Impact of training node budget on GPU memory usage by FALCON-CQSIGN (EC).}
\label{gpu_vs_budget}
\end{figure}

\begin{figure}[h!]
\centering
\includegraphics[width=1.0\columnwidth]{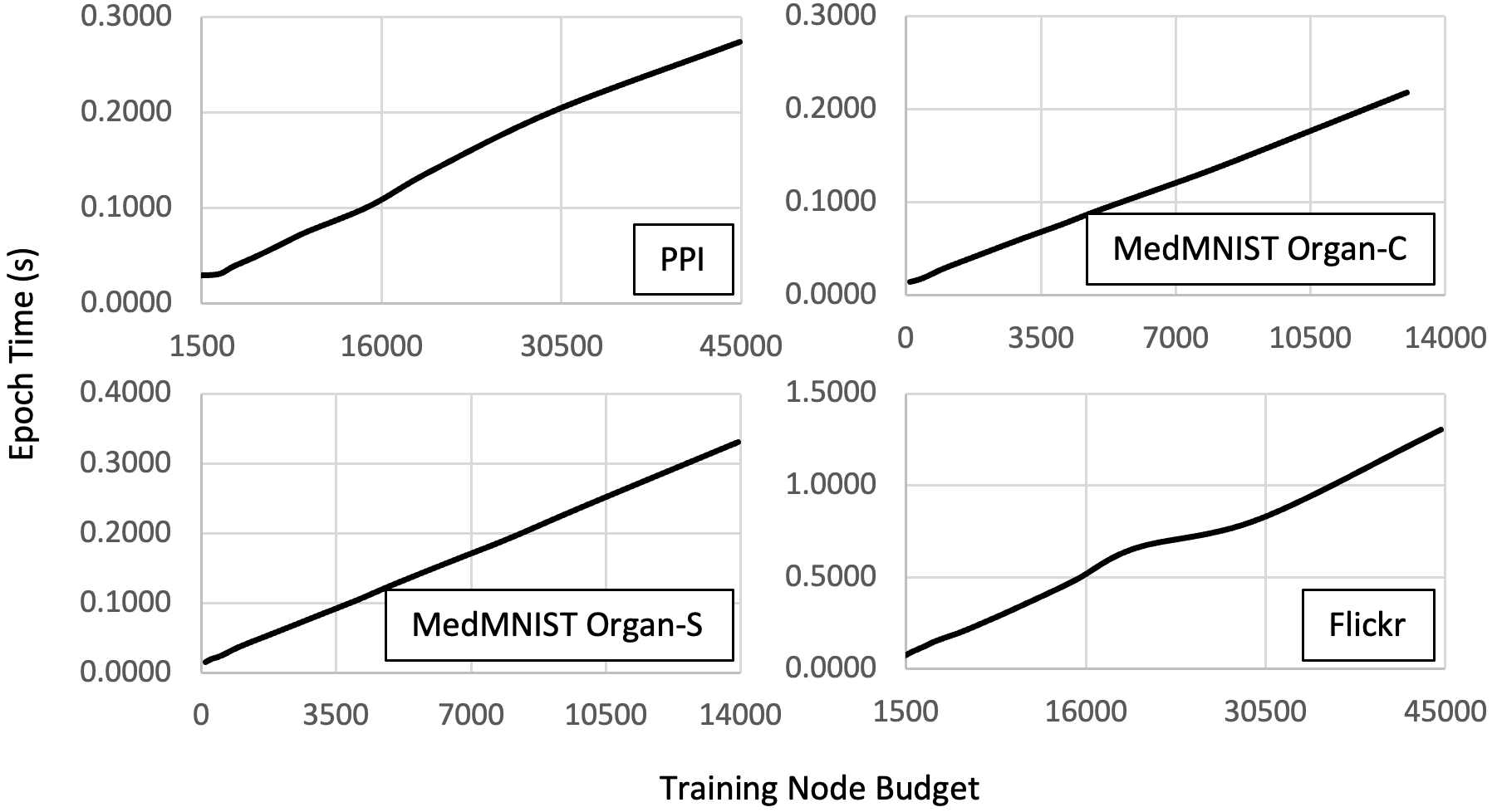}
\caption{Impact of training node budget on epoch time by FALCON-CQSIGN (EC).}
\label{time_vs_budget}
\end{figure}

In this section, we show the impact of the FALCON node budget on the GPU memory footprint during training. Figure~\ref{gpu_vs_budget} shows the GPU memory usage of the FALCON-CQSIGN (EC) framework. Similarly, for epoch time, we plot the epoch time against the training node budget in Figure~\ref{time_vs_budget}.

\section{Computational time of FALCON}
Here, we compare the computational time required to collapse the training graphs using FALCON. This computational time is shown in Table~\ref{tab:contraction_time} and consists of the whole collapse procedure (centrality measure computation, K-means clustering, and the graph collapsing process). Additionally, graph coarsening \cite{huang2021scaling} is compared with FALCON, which suggests that most FALCON methods are faster, excluding the CC variant.

\begin{table}[h!]
  \centering
  \caption{Computational time comparison of various FALCON methods and graph coarsening \cite{huang2021scaling} on various datasets.}
  \scalebox{0.85}{
    \begin{tabular}{@{}p{7.2em}|>{\centering\arraybackslash}p{2.8em}>{\centering\arraybackslash}p{2.8em}>{\centering\arraybackslash}p{2.8em}>{\centering\arraybackslash}p{2.8em}>{\centering\arraybackslash}p{2.8em}>{\centering\arraybackslash}p{2.8em}@{}}
    Collapsing Method & PPI & Flickr & Organ-S & Organ-C & Cora  & PubMed \\
    \midrule
    FALCON (DC) & \underline{\textbf{2.85}} & \underline{\textbf{9.59}} & \underline{\textbf{4.57}} & \underline{\textbf{4.49}} & \underline{\textbf{0.66}} & \underline{\textbf{1.90}} \\
    FALCON (BC) & \underline{13.78} & \underline{93.44} & 11.64 & 13.90 & 1.01 & 11.14 \\
    FALCON (PR) & \textbf{4.70} & \textbf{11.76} & \underline{5.81} & \underline{6.65} & \textbf{0.51} & \textbf{2.48} \\
    FALCON (CC) & 177.61 & -     & 316.24 & 487.25 & 5.33 & 413.09 \\
    FALCON (EC) & 66.48 & 104.24 & \textbf{5.74} & \textbf{6.25} & \underline{0.62} & \underline{3.42} \\
    Coarsen \cite{huang2021scaling} & -     & -     & -     & -     & 1.16 & 18.05 \\
    \end{tabular}%
    }
  \label{tab:contraction_time}%
\begin{minipage}{1\columnwidth}
\vspace{0.1cm}
\vspace{0.1cm}
\small  Notes: \underline{\textbf{Bold underline}}, \textbf{bold}, and  \underline{underline} denotes 1\textsuperscript{st}, 2\textsuperscript{nd}, and 3\textsuperscript{rd} best result of each models, respectively. Note that results are in seconds.
\end{minipage}
\end{table}%

\begin{table*}[h!]
  \centering
  \caption{More prediction result comparison of FALCON and Pure Centrality-Based Collapse \cite{cqsign}.}
  \scalebox{0.8}{
    \begin{tabular}{p{12em}|>{\centering\arraybackslash}p{4.2em}>{\centering\arraybackslash}p{4.2em}>{\centering\arraybackslash}p{4.2em}>{\centering\arraybackslash}p{4.2em}|>{\centering\arraybackslash}p{4.2em}>{\centering\arraybackslash}p{4.2em}|>{\centering\arraybackslash}p{4.2em}>{\centering\arraybackslash}p{4.2em}|>{\centering\arraybackslash}p{4.2em}>{\centering\arraybackslash}p{4.2em}}
    \multirow{2}[1]{*}{Methods} & \multicolumn{4}{c|}{PPI}      & \multicolumn{2}{c|}{MedMNIST Organ-S} & \multicolumn{2}{c|}{MedMNIST Organ-C} & \multicolumn{2}{c}{Flickr} \\
    \multicolumn{1}{c|}{} & ACC   & F1    & Sensitivity & Specificity & ACC   & Specificity & ACC   & Specificity & ACC   & Specificity \\
    \midrule
    FALCON-GCN (DC) & \textbf{82.17±0.91} & 68.67±1.72 & \textbf{65.30±1.87} & \textbf{89.38±0.51} & 58.80±0.56 & 95.88±0.06 & 74.61±0.48 & 97.46±0.05 & \textbf{48.07±0.54} & \textbf{91.34±0.09} \\
    FALCON-GCN (BC) & \textbf{88.98±0.33} & \textbf{80.37±0.69} & 75.38±1.04 & \textbf{94.79±0.07} & \textbf{59.04±0.90} & \textbf{95.90±0.09} & 75.15±0.51 & 97.52±0.05 & \textbf{47.66±0.56} & \textbf{91.28±0.09} \\
    FALCON-GCN (PR) & \textbf{81.93±0.52} & \textbf{66.40±1.45} & 59.68±2.13 & \textbf{91.44±0.34} & \textbf{58.99±0.41} & \textbf{95.90±0.04} & 74.76±0.34 & 97.48±0.03 & \textbf{47.95±0.54} & \textbf{91.33±0.09} \\
    FALCON-GCN (CC) & \textbf{97.69±0.12} & \textbf{96.10±0.20} & \textbf{94.93±0.29} & \textbf{98.87±0.08} & \textbf{58.99±0.43} & \textbf{95.90±0.04} & 74.57±0.62 & 97.46±0.06 & -     & - \\
    FALCON-GCN (EC) & \textbf{98.76±0.02} & \textbf{97.91±0.03} & \textbf{97.33±0.10} & 99.36±0.04 & \textbf{59.06±0.56} & \textbf{95.91±0.06} & 75.54±0.27 & 97.55±0.03 & \textbf{47.34±0.80} & \textbf{91.22±0.13} \\
    \midrule
    FALCON-GAS (DC) & \textbf{78.21±0.70} & \textbf{60.02±0.81} & \textbf{54.61±0.45} & \textbf{88.30±1.01} & \textbf{57.36±0.78} & \textbf{95.74±0.08} & \textbf{74.23±1.15} & \textbf{97.42±0.11} & 47.24±0.87 & \textbf{91.21±0.14} \\
    FALCON-GAS (BC) & \textbf{87.61±0.15} & \textbf{77.46±0.30} & 71.08±0.49 & \textbf{94.68±0.20} & \textbf{57.89±0.25} & \textbf{95.79±0.02} & 74.15±0.85 & 97.41±0.09 & \textbf{47.43±0.79} & 91.24±0.13 \\
    FALCON-GAS (PR) & \textbf{79.07±0.20} & \textbf{57.65±0.73} & 47.62±1.35 & \textbf{92.51±0.63} & \textbf{58.02±0.71} & \textbf{95.80±0.07} & \textbf{74.33±0.79} & \textbf{97.43±0.08} & \textbf{47.99±0.87} & \textbf{91.33±0.15} \\
    FALCON-GAS (CC) & \textbf{96.80±0.12} & \textbf{94.56±0.20} & \textbf{92.94±0.16} & 98.44±0.10 & \textbf{58.57±0.19} & \textbf{95.86±0.02} & \textbf{74.88±0.44} & \textbf{97.49±0.04} & -     & - \\
    FALCON-GAS (EC) & 98.41±0.10 & 97.32±0.17 & \textbf{96.79±0.27} & 99.10±0.09 & \textbf{58.20±0.63} & \textbf{95.82±0.06} & \textbf{73.97±0.85} & \textbf{97.40±0.08} & \textbf{47.51±0.35} & \textbf{91.25±0.06} \\
    \midrule
    FALCON-ClusterGCN (DC) & \textbf{84.66±0.08} & 71.60±0.15 & 64.61±0.29 & \textbf{93.23±0.14} & \textbf{59.53±0.75} & \textbf{95.95±0.08} & \textbf{76.10±1.65} & \textbf{97.61±0.16} & \textbf{46.39±0.14} & \textbf{91.07±0.02} \\
    FALCON-ClusterGCN (BC) & \textbf{90.12±0.14} & 82.64±0.27 & 78.55±0.38 & \textbf{95.06±0.07} & \textbf{60.43±0.87} & \textbf{96.04±0.09} & \textbf{77.16±1.63} & \textbf{97.72±0.16} & \textbf{46.33±0.12} & \textbf{91.05±0.02} \\
    FALCON-ClusterGCN (PR) & \textbf{84.01±0.09} & 70.25±0.21 & 63.06±0.40 & \textbf{92.96±0.16} & \textbf{59.25±0.94} & \textbf{95.92±0.09} & \textbf{76.65±1.20} & \textbf{97.67±0.12} & \textbf{46.40±0.15} & \textbf{91.07±0.02} \\
    FALCON-ClusterGCN (CC) & 96.65±0.28 & 94.34±0.49 & 93.21±0.61 & \textbf{98.12±0.14} & \textbf{59.59±0.66} & \textbf{95.96±0.07} & \textbf{76.36±1.16} & \textbf{97.64±0.12} & -     & - \\
    FALCON-ClusterGCN (EC) & 97.95±0.40 & 96.56±0.67 & 96.14±0.71 & 98.72±0.27 & \textbf{59.23±0.94} & \textbf{95.92±0.09} & \textbf{76.15±1.14} & \textbf{97.62±0.11} & \textbf{46.12±0.21} & \textbf{91.02±0.04} \\
    \midrule
    FALCON-SIGN (DC) & \textbf{80.93±0.05} & 61.43±0.31 & 50.72±0.60 & \textbf{93.84±0.24} & \textbf{60.82±0.49} & \textbf{96.08±0.05} & \textbf{77.95±0.47} & \textbf{97.80±0.05} & 48.43±0.62 & 91.40±0.10 \\
    FALCON-SIGN (BC) & \textbf{87.71±0.05} & 77.23±0.10 & 69.62±0.14 & \textbf{95.43±0.05} & \textbf{60.51±0.61} & \textbf{96.05±0.06} & \textbf{78.01±0.54} & \textbf{97.80±0.05} & \textbf{49.05±0.41} & \textbf{91.51±0.07} \\
    FALCON-SIGN (PR) & \textbf{80.31±0.06} & 59.16±0.17 & 47.63±0.34 & \textbf{94.27±0.18} & \textbf{60.59±0.40} & \textbf{96.06±0.04} & \textbf{77.93±0.37} & \textbf{97.79±0.04} & 48.58±0.63 & 91.43±0.10 \\
    FALCON-SIGN (CC) & \textbf{97.39±0.02} & \textbf{95.56±0.03} & \textbf{93.66±0.07} & \textbf{98.98±0.03} & \textbf{60.68±0.18} & \textbf{96.07±0.02} & 77.53±0.41 & 97.75±0.04 & -     & - \\
    FALCON-SIGN (EC) & \textbf{98.68±0.04} & \textbf{97.78±0.06} & \textbf{97.02±0.06} & 99.39±0.05 & \textbf{60.59±0.29} & \textbf{96.06±0.03} & \textbf{77.81±0.64} & \textbf{97.78±0.06} & \textbf{47.92±0.46} & \textbf{91.32±0.08} \\
    \midrule
    \midrule
    Collapse-GCN (DC) & 77.93±0.18 & \textbf{63.45±0.56} & 64.00±1.37 & 83.88±0.58 & \textbf{58.81±0.50} & \textbf{95.88±0.05} & \textbf{76.10±0.37} & \textbf{97.61±0.04} & 48.00±0.36 & 91.33±0.06 \\
    Collapse-GCN (BC) & 87.55±0.07 & 78.48±0.16 & \textbf{75.83±0.33} & 92.56±0.10 & 58.75±0.45 & 95.88±0.05 & \textbf{76.08±0.47} & \textbf{97.61±0.05} & 47.29±0.34 & 91.21±0.06 \\
    Collapse-GCN (PR) & 79.08±0.04 & 64.93±0.15 & \textbf{64.68±0.33} & 85.23±0.11 & 58.81±0.50 & 95.88±0.05 & \textbf{76.10±0.37} & \textbf{97.61±0.04} & 47.38±0.47 & 91.23±0.08 \\
    Collapse-GCN (CC) & 97.19±0.03 & 95.24±0.03 & 94.03±0.20 & \textbf{98.53±0.12} & 58.81±0.50 & 95.88±0.05 & \textbf{76.10±0.37} & \textbf{97.61±0.04} & -     & - \\
    Collapse-GCN (EC) & 98.72±0.01 & 97.85±0.02 & 97.10±0.05 & \textbf{99.41±0.01} & 58.81±0.50 & 95.88±0.05 & \textbf{76.10±0.37} & \textbf{97.61±0.04} & 46.77±0.31 & 91.13±0.05 \\
    \midrule
    Collapse-GAS (DC) & 73.83±0.49 & 53.01±2.84 & 49.67±5.84 & 84.16±2.84 & 56.88±0.32 & 95.69±0.03 & 73.80±1.18 & 97.38±0.12 & \textbf{47.27±1.38} & 91.21±0.23 \\
    Collapse-GAS (BC) & 86.82±0.09 & 76.99±0.25 & \textbf{73.66±0.68} & 92.45±0.24 & 57.15±0.51 & 95.72±0.05 & \textbf{74.85±0.61} & \textbf{97.49±0.06} & 47.43±0.80 & 91.24±0.13 \\
    Collapse-GAS (PR) & 76.44±0.16 & 57.30±0.63 & \textbf{52.81±1.26} & 86.54±0.58 & 56.88±0.32 & 95.69±0.03 & 73.80±1.18 & 97.38±0.12 & 47.42±1.08 & 91.24±0.18 \\
    Collapse-GAS (CC) & 96.28±0.12 & 93.70±0.23 & 92.44±0.55 & 97.92±0.11 & 56.88±0.32 & 95.69±0.03 & 73.80±1.18 & 97.38±0.12 & -     & - \\
    Collapse-GAS (EC) & \textbf{98.41±0.06} & \textbf{97.33±0.11} & 96.58±0.15 & 99.20±0.06 & 56.88±0.32 & 95.69±0.03 & 73.80±1.18 & 97.38±0.12 & 46.77±0.75 & 91.13±0.13 \\
    \midrule
    Collapse-ClusterGCN (DC) & 83.31±0.09 & \textbf{72.63±0.22} & \textbf{73.99±0.55} & 87.29±0.21 & 58.95±0.52 & 95.90±0.05 & 75.84±0.72 & 97.58±0.07 & 45.85±0.16 & 90.98±0.03 \\
    Collapse-ClusterGCN (BC) & 89.82±0.16 & \textbf{83.00±0.24} & \textbf{83.02±0.27} & 92.73±0.21 & 59.45±0.43 & 95.95±0.04 & 76.42±0.40 & 97.64±0.04 & 45.77±0.16 & 90.96±0.03 \\
    Collapse-ClusterGCN (PR) & 82.67±0.09 & \textbf{71.78±0.19} & \textbf{73.63±0.55} & 86.53±0.26 & 59.06±0.67 & 95.91±0.07 & 76.10±0.38 & 97.61±0.04 & 46.00±0.14 & 91.00±0.02 \\
    Collapse-ClusterGCN (CC) & \textbf{96.69±0.02} & \textbf{94.44±0.03} & \textbf{93.88±0.10} & 97.89±0.05 & 58.71±0.26 & 95.87±0.03 & 75.25±1.24 & 97.52±0.12 & -     & - \\
    Collapse-ClusterGCN (EC) & \textbf{98.10±0.03} & \textbf{96.81±0.04} & \textbf{96.38±0.11} & \textbf{98.84±0.05} & 58.81±0.42 & 95.88±0.04 & 75.36±0.94 & 97.54±0.09 & 45.97±0.10 & 91.00±0.02 \\
    \midrule
    Collapse-SIGN (DC) & 80.55±0.08 & \textbf{66.37±0.06} & \textbf{64.11±0.26} & 87.58±0.22 & 60.44±0.19 & 96.04±0.02 & 77.61±0.15 & 97.76±0.01 & \textbf{48.54±0.49} & \textbf{91.42±0.08} \\
    Collapse-SIGN (BC) & 87.47±0.05 & \textbf{77.73±0.05} & \textbf{73.03±0.32} & 93.64±0.19 & 60.01±0.23 & 96.00±0.02 & 77.86±0.38 & 97.79±0.04 & 48.33±0.59 & 91.39±0.10 \\
    Collapse-SIGN (PR) & 80.19±0.07 & \textbf{64.44±0.17} & \textbf{59.97±0.24} & 88.83±0.03 & 60.44±0.19 & 96.04±0.02 & 77.61±0.15 & 97.76±0.01 & \textbf{48.93±0.54} & \textbf{91.49±0.09} \\
    Collapse-SIGN (CC) & 97.17±0.03 & 95.20±0.05 & 93.65±0.02 & 98.67±0.04 & 60.44±0.19 & 96.04±0.02 & \textbf{77.61±0.15} & \textbf{97.76±0.01} & -     & - \\
    Collapse-SIGN (EC) & 98.66±0.01 & 97.75±0.02 & 96.94±0.07 & \textbf{99.40±0.02} & 60.44±0.19 & 96.04±0.02 & 77.61±0.15 & 97.76±0.01 & 47.92±0.56 & 91.32±0.09 \\
    \end{tabular}%
    }
  \label{tab:moreFLP_vsNonFLP}%
\begin{minipage}{2\columnwidth}
\vspace{0.1cm}
\vspace{0.1cm}
\small  Notes: Results are \textbf{bolded} if they are better than their counterparts.
\end{minipage}
\end{table*}%

\section{More Comparison on Feature-Label Constraint Impact}
This section includes more comparisons between FALCON and its feature-label agnostic counterparts (Table~\ref{tab:moreFLP_vsNonFLP}). Our experiment suggests that the preservation of the original feature and label distributions generally tends to be beneficial in terms of prediction quality across various GNN models as well as various datasets.

\end{appendices}

\bibliographystyle{IEEEtran}
\bibliography{citations}

\end{document}